\documentclass{article}

\usepackage{colm2024_conference}

\usepackage{microtype}
\usepackage{hyperref}
\usepackage{url}
\usepackage{booktabs}

\usepackage{graphicx}
\usepackage{amsmath}
\usepackage{amssymb}
\usepackage{multirow}
\usepackage{wrapfig}
\usepackage{enumitem}

\usepackage[accsupp]{axessibility}  

\colmfinalcopy

\usepackage[capitalize]{cleveref}
\crefname{section}{Sec.}{Secs.}
\Crefname{section}{Section}{Sections}
\Crefname{table}{Table}{Tables}
\crefname{table}{Tab.}{Tabs.}

\definecolor{lightblue}{rgb}{0.90, 0.95, 0.98}

\newcommand{\bfit}[1]{\textbf{\textit{#1}}}

\newcommand{\sdonefour}{Stable Diffusion v1.4}

\newcommand{\vpgen}{VPGen}
\newcommand{\autotikz}{AutomaTikZ}
\newcommand{\method}{DiagrammerGPT}
\newcommand{\generator}{DiagramGLIGEN}
\newcommand{\dataset}{AI2D-Caption}
\newcommand{\auditor}{auditor}

\newcommand{\plan}{diagram plan}

\newcommand{\textpackage}{Pillow}

\newcommand{\eg}[1]{\emph{e.g}.}
\newcommand{\ie}[1]{\emph{i.e}.}
\newcommand{\etc}[1]{\emph{etc}.}
\newcommand{\vs}[1]{\emph{vs}.}
\newcommand{\etal}[1]{\emph{et al}.}

\definecolor{Red}{rgb}{0.6,0,0}
\definecolor{Blue}{rgb}{0,0,0.8}
\definecolor{Green}{rgb}{0.2,0.8,0}
\definecolor{airforceblue}{rgb}{0.36, 0.54, 0.66}
\definecolor{ao(english)}{rgb}{0.0, 0.5, 0.0}
\definecolor{azure(colorwheel)}{rgb}{0.0, 0.5, 1.0}
\definecolor{crimson}{rgb}{0.86, 0.08, 0.24}
\definecolor{darkcerulean}{rgb}{0.03, 0.27, 0.49}
\definecolor{cobalt}{rgb}{0.0, 0.28, 0.67}
\definecolor{rosegold}{rgb}{0.72, 0.43, 0.47}
\definecolor{orange-red}{rgb}{1.0, 0.27, 0.0}
\definecolor{mountainmeadow}{rgb}{0.19, 0.73, 0.56}
\definecolor{malachite}{rgb}{0.04, 0.85, 0.32}
\definecolor{darkblue}{rgb}{0.0, 0.0, 0.55}
\definecolor{customblue}{rgb}{0.2, 0.35, 0.8}

\hypersetup{colorlinks,linkcolor={blue},citecolor={mountainmeadow},urlcolor={magenta}}

\title{\method{}: Generating Open-Domain, Open-Platform Diagrams via LLM Planning}
\author{Abhay Zala \qquad
Han Lin \qquad
Jaemin Cho \qquad
Mohit Bansal \\
UNC Chapel Hill\\
\texttt{\{aszala, hanlincs, jmincho, mbansal\}@cs.unc.edu} \\
\\
\textbf{\url{https://diagrammerGPT.github.io}}
}

\begin{document}

\maketitle

\begin{abstract}
Text-to-image (T2I) generation has seen significant growth
over the past few years.
Despite this, there has been little work on generating diagrams with T2I models.
A diagram is a symbolic/schematic representation that explains information using structurally rich and spatially complex visualizations (\eg{}, a dense combination of related objects, text labels, directional arrows/lines, \etc{}).
Existing state-of-the-art T2I models often fail at diagram generation because they lack fine-grained object layout control when many objects are densely connected via complex relations such as arrows/lines, and also often fail to render comprehensible text labels.
To address this gap, we present \method{}, a novel two-stage text-to-diagram generation framework leveraging the layout guidance capabilities of LLMs to generate more accurate diagrams.
In the first stage,
we use LLMs to generate and iteratively refine
`\plan{}s' (in a planner-\auditor{} feedback loop).
In the second stage,
we use a diagram generator, \generator{},
and a text label rendering module to generate diagrams (with clear text labels) following the \plan{}s.
To benchmark the text-to-diagram generation task, we introduce \dataset{}, a densely annotated diagram dataset built on top of the AI2D dataset.
We show that our \method{} framework produces more accurate diagrams, outperforming existing T2I models.
We also provide comprehensive analysis, including
open-domain diagram generation,
multi-platform vector graphic diagram generation,
human-in-the-loop editing, and
multimodal planner/\auditor{} LLMs.
\end{abstract}

\section{Introduction}
\label{sec:intro}

Over the past few years,
text-to-image (T2I) generation models~\citep{stablediffusion,Ramesh2022UnCLIP,Yu2022Parti,Chang2023Muse,dai2023emu}
have shown impressive advancements in image generation quality.
Large language models (LLMs) have also recently shown strong capabilities and usefulness in broad language understanding and generation tasks~\citep{Touvron2023LLaMA1,Touvron2023Llama2,OpenAI2023GPT4TR,Chung2022ScalingIL,Brown2020GPT3,Chowdhery2022PaLMSL}.
Recent works also have demonstrated that it is possible to leverage LLMs to control layouts for the downstream T2I models, for better semantic alignment with input text prompts~\citep{Cho2023VPT2I,feng2023layoutgpt,lian2023llmgrounded}.

However, it has not been explored to use the combination of LLM and T2I generation frameworks for creating diagrams.
A diagram is a symbolic/schematic representation that explains information using rich and spatially complex visualizations (\eg{}, a dense combination of objects, text labels, arrows, lines, \etc{}).
A system that helps to create accurate diagrams would be useful for preparing many educational and academic resources (\eg{}, creating a diagram that explains new concepts in books, presentations, and papers).
While the existing T2I generation models are good at generating realistic images,
they often fail at diagram generation because they lack fine-grained object layout control when many objects are densely connected via complex relations such as arrows/lines and also often fail to render comprehensible text labels.
In diagram generation, it is more important to convey correct information (with correct object relationships) than to generate photorealistic objects.
In our experiments,
existing T2I generation models
usually generate diagrams where objects and arrows/lines have incorrect relationships and rendered text labels are incomprehensible.
Even the recent state-of-the-art (SOTA) model DALL-E 3~\citep{OpenAI2023DALLE3} struggles to render accurate diagrams (mention in their system card and shown in \cref{fig:open_domain_generation_half}).

To address the issues in text-to-diagram generation,
we introduce \method{}, a novel two-stage framework capable of generating more accurate open-domain diagrams for multiple platforms by following LLM-generated \plan{}s.
As shown in \cref{fig:teaser},
our \method{} splits the text-to-diagram generation task into two stages: \textbf{diagram planning} and \textbf{diagram generation}.
For the first stage (\cref{fig:teaser} left),
we employ an LLM to create and refine `\textit{\plan{}s}'.
In the second stage (\cref{fig:teaser} right),
we introduce \generator{}, a layout-guided diagram generation module,
to generate diagrams based on \plan{}s, then explicitly render text labels, ensuring their readability.

{
\begin{figure}
    \centering
    \includegraphics[width=0.99\textwidth]{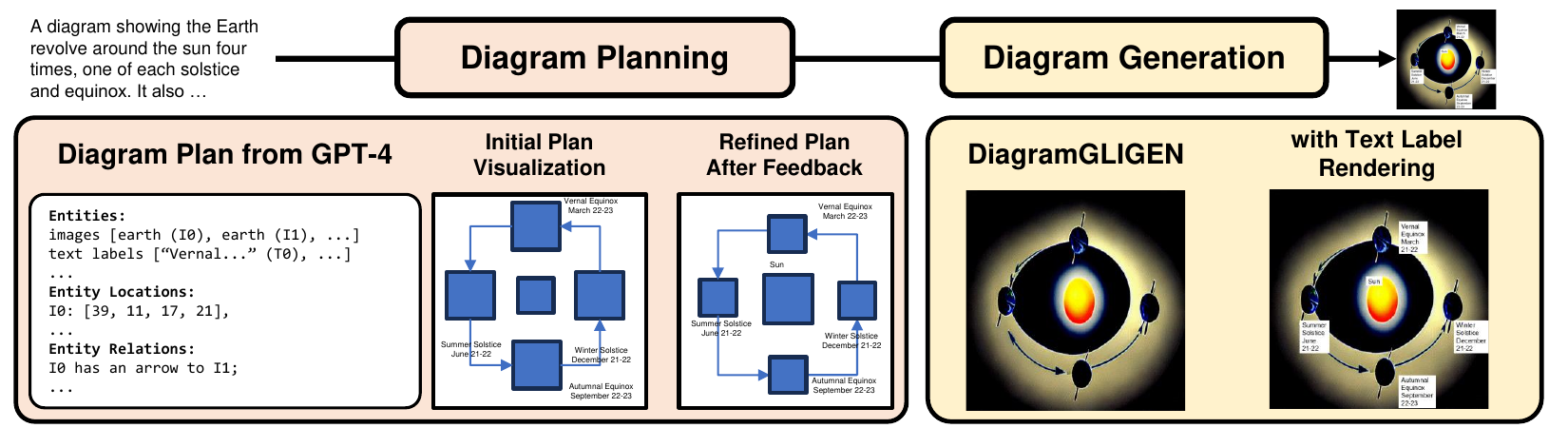}
    \caption{
    An overview of \method{}, our two-stage framework for open-domain diagram generation.
    In the first diagram planning stage (\cref{subsec:stage1_method}),
    given a prompt, our LLM (GPT-4~\citep{OpenAI2023GPT4TR}) generates a \textit{\plan{}}, which consists of dense entities, fine-grained relationships, and precise layouts.
    Then, the LLM iteratively refines the plan to correct mistakes.
    In the second diagram generation stage (\cref{subsec:stage2_method}),
    our \generator{}  outputs the diagram given the \plan{},
    then, we render the text labels on the diagram.
    }
    \label{fig:teaser}
\end{figure}
}

In the first stage, diagram planning (\cref{subsec:stage1_method}), we employ
an LLM (\eg{}, GPT-4~\citep{OpenAI2023GPT4TR}) to act as a \textit{planner} and generate \textbf{\plan{}s} given text prompts.
The \plan{}s consist of
(1) a list of entities (\ie{}, objects and text labels),
(2) relationships between the entities (\ie{}, arrows or lines between entities),
and (3) entity layouts (\ie{}, 2D bounding box coordinates).
After the initial generation of \plan{}s,
inspired by recent works on LLMs that self-refine previously generated contents~\citep{Chen2023TeachingLL,Miao2023SelfCheckUL,Madaan2023SelfRefineIR},
we use another LLM to act as an \textit{\auditor{}} and find potential errors such as incorrect object positions or relationships between objects.
Then, the planner LLM takes feedback from the \auditor{} LLM to update the \plan{} (\eg{}, in \cref{fig:teaser}, the sun is smaller than the earths, so the plan is adjusted to fix the relative scales).
We find that our LLM-generated \plan{}s are quite accurate and that refinement can effectively help correct some small errors (see \cref{subsec:human_eval_results} and feedback example in appendix for more details).

In the second stage, diagram generation (\cref{subsec:stage2_method}), we introduce \generator{}, a layout-guided diagram generation module,
to generate diagrams based on \plan{}s,
then explicitly render text labels, ensuring their readability.
We implement \generator{} based on GLIGEN~\citep{li2023gligen} architecture, which adds gated self-attention layers to the \sdonefour{}~\citep{stablediffusion} model.
While the original GLIGEN model is only trained on natural images and
take only objects for layout grounding,
\generator{}
is more specialized in the diagram domain,
by
being trained on our new \dataset{} diagram dataset (see the following paragraph and \cref{subsec:dataset_collection} for more details)
and taking the text labels and arrows as additional layout grounding inputs.
As the SOTA diffusion models still struggle in rendering text~\citep{liu-etal-2023-character,Chen2023TextDiffuserDM,Cho2023VPT2I},
we explicitly render text labels on top of the generated diagrams to ensure they are readable.
Compared to the T2I baselines,
our diagram generation stage allows
more accurate object layouts and relationships between the objects (\eg{}, arrows/lines), and clear text labels.

Since there are no diagram datasets with detailed captions and fine-grained layout annotations,
we construct \textbf{\dataset{}}, a new dataset for the text-to-diagram generation task built on top of the AI2D~\citep{AI2D} dataset (\cref{subsec:dataset_collection}).
We employ LLaVA 1.5~\citep{liu2023improvedLLava}, a SOTA multimodal LLM, to create annotations of captions and object descriptions on AI2D diagrams.

We comprehensively compare our \method{} to
recent
text-to-image/diagram generation methods, including
\sdonefour{}~\citep{stablediffusion},
\vpgen{}~\citep{Cho2023VPT2I},
and \autotikz{}~\citep{Belouadi2023AutomaTikZTS},
in both zeroshot and fine-tuned settings (see \cref{subsec:baseline_models} and \cref{subsec:main_results}).
In our quantitative and qualitative analysis,
our \method{} demonstrates more accurate diagram generation performance than the baseline models.
Our method also outperforms \sdonefour{}, the closest and strongest baseline, in a human preference study on both image-text alignment and object relationship criteria.
In our error analysis, we show that \plan{}s generated by our LLM (GPT-4) are quite accurate (before and after refinement), while \generator{} sometimes makes mistakes when generating diagrams.
This indicates that our \method{} can benefit from future layout-guided image generation backbones stronger than GLIGEN (see \cref{subsec:human_eval_results}).

\begin{figure}[t]
    \centering
    \includegraphics[width=\textwidth]{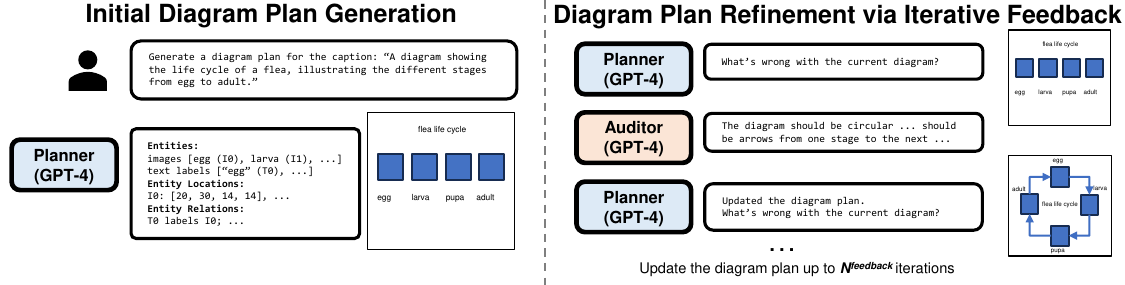}
    \caption{
    Illustration of the first stage of \method{}: diagram planning (\cref{subsec:stage1_method}).
    We use a planner LLM (\eg{}, GPT-4~\citep{OpenAI2023GPT4TR}) to create the fine-grained layouts of diagrams, which we call \textit{\plan{}s}.
    We first generate an initial diagram from the input text prompt with an LLM (\textbf{left}). Then we iteratively refine \plan{}s in a feedback loop of the planner and \auditor{} LLMs.
    }
    \label{fig:diagram_plan}
\end{figure}

We also conduct additional analysis, including
\textbf{open-domain} diagram generation,
\textbf{multiple platforms vector graphic} diagram generation,
\textbf{human-in-the-loop} editing,
and \textbf{text-only \vs{} multimodal LLM} (\eg{}, GPT-4Vision~\citep{OpenAI2023GPT4Vison}; see \cref{subsec:additional_analysis_results} and appendix).
First, we experiment with generating diagrams in unseen domains (\eg{}, geology and plants) that are not covered in the LLM in-context learning domains (astronomy, biology, engineering),
where our \method{} could often generate semantically accurate \plan{}s and diagrams from these unseen prompts.
Second, we experiment with rendering our \plan{}s in multiple platforms:
Microsoft PowerPoint, Inkscape, and Adobe Illustrator (see \cref{fig:vector_diagram_examples_half}).
Third, we show that once a \plan{} is exported to another platform, end-users can also manually edit plans to their liking and send them back to our \generator{} to generate images using their manually refined plan (\ie{}, human-in-the-loop refinement of \plan{}).
Lastly, we experiment with using the recently released GPT-4Vision~\citep{OpenAI2023GPT4Vison} instead of text-only GPT-4
for our planner and \auditor{} LLMs (in appendix).

\section{Related Works}
\label{sec:rel_works}

\subsection{Text-to-Image Generation}
\label{sec:rel_works_t2i}
In the text-to-image (T2I) generation task, models generate an image from a given text prompt.
Recently multimodal language models (\eg{}, Parti~\citep{Yu2022Parti} and MUSE~\citep{Chang2023Muse})
and diffusion models (\eg{}, Stable Diffusion~\citep{stablediffusion}, DALL-E 2~\citep{Ramesh2022UnCLIP},
and Imagen~\citep{Saharia2022Imagen}) have gained popularity for this task.
These recent T2I generation models have demonstrated impressive photorealism in their zeroshot image generation capabilities.
However, using these models directly for diagram generation is challenging due to the significant distribution gap between diagrams and their training images.
Additionally, these models often lack precise control of fine-grained layouts when many objects are densely connected via complex relations and frequently produce illegible text labels, which are essential for the diagram generation task.

\subsection{Text-to-Image Generation with LLM-Guided Layouts}
Large language models (LLMs)
have demonstrated their usefulness in various language generation tasks \citep{Touvron2023LLaMA1,Touvron2023Llama2,OpenAI2023GPT4TR,Chung2022ScalingIL,Brown2020GPT3,Chowdhery2022PaLMSL}.
Recent works also leverage LLMs to control layouts for the downstream image generation models, for better semantic alignment with input text prompts~\citep{Cho2023VPT2I,feng2023layoutgpt,lian2023llmgrounded}.
However, these works focus on generating natural images and do not have capabilities that are crucial in diagram generations, \eg{}, rendering clear text labels or the ability to precisely control fine-grained layouts of many objects that are densely connected via complex relations such as arrows/lines, as shown in our experiments (\cref{sec:results}).
A concurrent work, \autotikz{}~\citep{Belouadi2023AutomaTikZTS} uses LLMs to generate
TikZ~\citep{Tikz}\footnote{A TeX package for generating graphics by composing primitive objects with basic polygons.} code to produce scientific vector graphics.
While TikZ can be used to draw specific types of diagrams, such as 2D bar plots or directional acyclic graphs, it is difficult to generate diagrams including entities not supported by the TikZ primitives, such as animals.
In contrast, our \method{} allows for generating diagrams including diverse entities
(\eg{}, different animals in life cycle diagrams or different planets/stars in astronomy diagrams).

{
\begin{figure}
    \centering
    \includegraphics[width=\textwidth]{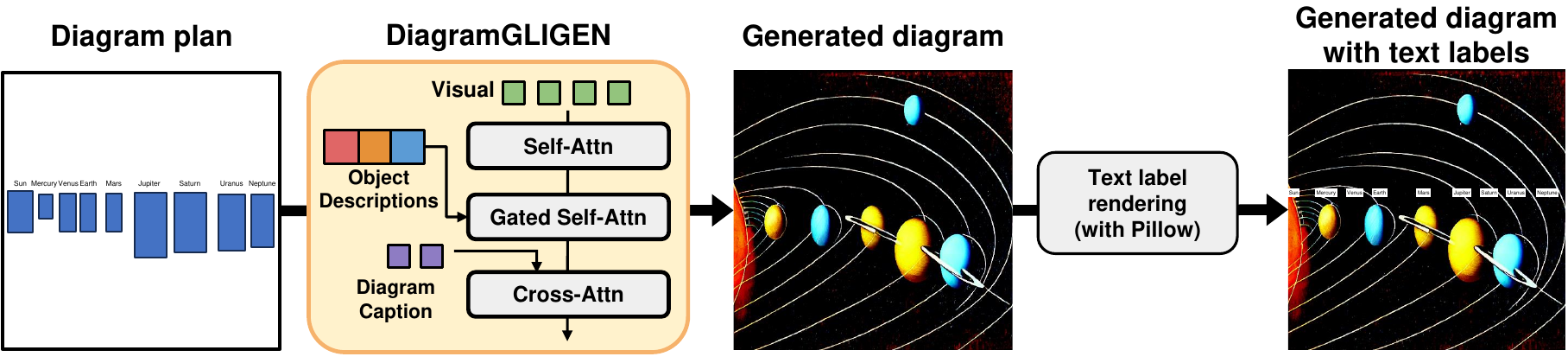}
    \caption{
    Illustration of the second stage of \method{}: diagram generation (\cref{subsec:stage2_method}). We first generate the objects from the \plan{} with \generator{}, our layout-guided diagram generation model. Then, we use \textpackage{} to render clear text labels.}
    \label{fig:diagram_generation_pipeline}
\end{figure}
}

\section{\method{}: Method Details}
\label{sec:method}
We introduce \method{}, a novel two-stage framework for generating open-domain diagrams from text prompts, where an LLM first generates
the overall plan, and a visual generator renders an actual diagram following the plan.
In the first stage, \textbf{Diagram Planning} (\cref{subsec:stage1_method}),
a \textit{planner} LLM
takes a text description of a diagram as input and generates a \textit{\plan{}}, an overall diagram layout that guides the downstream diagram generation module.
The \textit{planner} LLM generates the initial \plan{} and iteratively refines the \plan{} via feedback from an \auditor{} LLM.
In the second \textbf{Diagram Generation} stage (\cref{subsec:stage2_method}),
\generator{}, our new layout-guided diagram generation module, takes the \plan{} and generates the diagram. 
Finally, we render the text labels (which are from the \plan{}) onto the diagram to ensure clear and easy-to-read labels for each entity.

\subsection{Stage 1: Diagram Planning}
\label{subsec:stage1_method}
As illustrated in the left part of \cref{fig:diagram_plan},
we use an LLM
(e.g., GPT-4~\citep{OpenAI2023GPT4TR}) to create the overall layouts for diagrams, which we call \textit{\plan{}s} (see appendix for detailed plan configuration and example).

\noindent\textbf{Initial \plan{} generation.} 
We generate \plan{}s with GPT-4 via 10 in-context learning examples. Plans consist of three components:
(1) entities - a dense list of objects; (2) relationships - complex relationships between entities; and (3) layouts - 2D bounding boxes of the entities.
See appendix for \plan{} generation prompts/extra details.

\noindent\textbf{Diagram plan refinement via iterative feedback.}
Although the \textit{planner} (GPT-4) generates fairly accurate initial \plan{}s, we can further refine it to account for potentially missing or improperly arranged entities.
To address this issue, we introduce an \textit{\auditor{}} LLM that checks for any mismatch between the current \plan{} and the input prompt. It then provides feedback, enabling the \textit{planner} LLM to refine the \plan{}s.
Our \textit{\auditor{}} and \textit{planner} LLMs form a feedback loop to iteratively refine the \plan{}s.
For this \textit{\auditor{}} LLM, we employ GPT-4 again but with a different preamble and in-context examples designed to give useful feedback (see appendix for the prompts to initialize the \textit{\auditor{}} LLM).
We repeat the feedback loop for up to $N$ iterations, using $N$= 4 in our experiments.
The right part of \cref{fig:diagram_plan} exemplifies the feedback process.

\subsection{Stage 2: Diagram Generation}
\label{subsec:stage2_method}
As shown in \cref{fig:diagram_generation_pipeline},
we first generate the diagram images
following the \plan{} with \generator{},
then render text labels
on the diagram.

\noindent\textbf{\generator{}: Layout-guided diagram generation.}
Conveying factual information is more crucial than drawing photorealistic objects in diagram generation.
In our experiments,
we observe that existing T2I models often omit important objects, and generate incorrect object relationships and unreadable text labels (see \cref{sec:results} and \cref{fig:qualitative_examples_half}).
Therefore, we introduce \generator{}, a layout-guided text-to-diagram generation model to tackle these issues.
Inspired by GLIGEN~\citep{li2023gligen},
we implement \generator{} by incorporating gated self-attention layers, which take layout grounding inputs, into \sdonefour{}.
Furthermore, we enhance layout control by incorporating text labels and relationships as part of the layout grounding inputs during training, which can reduce the generation of unreadable text and redundant arrows/lines during inference (by not giving text labels during inference).
We employ CLIP~\citep{radford2021learning} text encoder to encode object descriptions and their relationships.
We use the CLIP image encoder to represent bounding box regions of the text labels in the ground truth diagrams. See the appendix for additional GLIGEN details and training setup.

\noindent\textbf{Text label rendering.}
Text labels (\eg{}, ``Sun'' labeling the sun object in \cref{fig:diagram_generation_pipeline})
in diagrams can effectively assist readers in understanding new concepts~\citep{circuit_learning_diagram_label_importance}.
However, as shown in Cho~\etal{}~\citep{Cho2023VPT2I}, existing T2I generation models (including \sdonefour{}) still struggle to generate high-quality text labels.
Therefore,
we explicitly render clear text labels on the diagrams with the \textpackage{}
Python package~\citep{clark2015pillow}.

\section{Experimental Setup}
\label{sec:exp_setups}

In the following subsections, we introduce our \dataset{} dataset for the text-to-diagram generation task (\cref{subsec:dataset_collection}),
baseline models (\cref{subsec:baseline_models}), evaluation metrics (\cref{subsec:metrics}),
and human evaluation setups (\cref{subsec:human_eval_setup}).

\begin{figure}[t]
    \centering
    \includegraphics[width=.8\columnwidth]{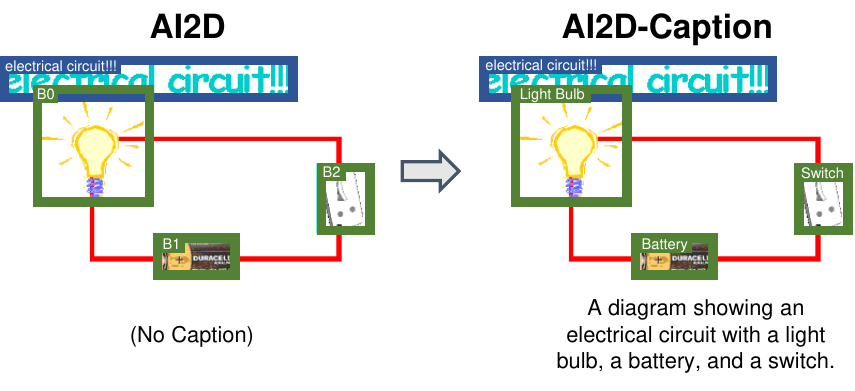}
    \caption{Example diagram annotation from the AI2D dataset~\citep{AI2D} (left) and our \dataset{} (right).
    \dataset{} additionally provides annotations of the diagram caption and bounding box region descriptions.
    }
    \label{fig:diagram_dataset}
\end{figure}

\subsection{\dataset{} Dataset}
\label{subsec:dataset_collection}

We introduce the \dataset{} dataset for the text-to-diagram generation task.
\dataset{} is built on top of AI2 Diagrams (AI2D) dataset~\citep{AI2D}, which provides annotations of around 4.9K diagrams covering diverse scientific domains, from Grade 1-6 science textbooks.
The original AI2D dataset's annotations are very short / are missing some aspects (\ie{}, each object bounding box is labeled simply as `blob'), so we employ LLaVA 1.5~\citep{liu2023improvedLLava}, a SOTA multimodal language model, to generate detailed captions for the diagrams and region descriptions of each bounding box (see \cref{fig:diagram_dataset} for comparison of AI2D and \dataset{} and the appendix for implementation details).
We use all of the LLaVA 1.5-aided caption and region annotations for \generator{} training and baseline fine-tuning.

To ensure that the LLM in-context examples and evaluation are accurate,
we manually annotate \plan{}s (\ie{}, captions, object/text label bounding boxes, object descriptions, arrows) for randomly selected 105 diagrams (held out from the training set described above).
Among the 105 diagrams, we use 30 diagrams that cover diverse scientific domains (10 for astronomy, 10 for biology, and 10 for engineering) as in-context examples, and 75 diagrams (25 for astronomy, 25 for biology, and 25 for engineering) as a test split.

\subsection{Baseline Models}
\label{subsec:baseline_models}
We compare our \method{} to several baseline models, including \sdonefour{}~\citep{stablediffusion},
\vpgen{} (Vicuna13B~\citep{vicuna2023} + GLIGEN)~\citep{Cho2023VPT2I},
and \autotikz{}~\citep{Belouadi2023AutomaTikZTS} (\textsc{CLiMA}-13B).
For the baselines, we experiment with zero-shot and fine-tuned (on our AI2D-Caption dataset where applicable) diagram generation.
For VPGen, we fine-tune both the Vicuna13B and GLIGEN.

\subsection{Evaluation Metrics}
\label{subsec:metrics}
\noindent \textbf{VPEval (objects, counts, relationships, texts).}
The VPEval metric~\citep{Cho2023VPT2I} evaluates diagrams in terms of the presence of objects (object evaluation), the number of objects (count evaluation), the correctness of spatial and connection relationships (relationship evaluation), and the presence of correct text labels (text evaluation).
While the original VPEval uses BLIP-2~\citep{Li2023BLIP2BL} as its visual reasoning model,
we employ the recently released
LLaVA 1.5~\citep{liu2023improvedLLava} model as our visual reasoning model, as we find it is more faithful in diagram question answering in our initial experiments and is an overall stronger model than BLIP-2~\citep{liu2023improvedLLava}.
See appendix for setup details.

\noindent\textbf{Captioning.}
In line with previous works~\citep{Hong2018,Hinz2020,Cho2022DallEval}, we also use captioning as a way to determine the accuracy of the generated diagrams.
We use LLaVA 1.5~\citep{liu2023improvedLLava} to caption each generated diagram and then compare this generated caption with the ground-truth caption using CIDEr~\citep{Vedantam2014CIDErCI} and BERTScore~\citep{Zhang20bertscore}.

\noindent\textbf{CLIPScore.}
Following previous works~\citep{Cho2022DallEval,Saharia2022Imagen,Belouadi2023AutomaTikZTS},
we use CLIPScore~\citep{hessel2021clipscore} to measure the similarity between the generated diagram and the original caption/ground-truth diagram.
Concretely, we calculate two types of CLIPScore:
(1) CLIPScore$^\text{Img-Txt}$: similarity between the generated diagram and the ground-truth caption.
(2) CLIPScore$^\text{Img-Img}$: similarity between the generated diagram and the ground-truth diagram.
We use OpenAI CLIP ViT-L/14~\citep{radford2021learning}.

\subsection{Human Evaluation}
\label{subsec:human_eval_setup}

\noindent
\textbf{Human error analysis of the two stages.}
As our \method{} pipeline consists of two stages, it is important to understand where any errors in our pipeline come from. 
In both stages, we assess 
(1) \textit{Object Presence}: whether all required objects for the diagrams are present, and
(2) \textit{Object Relationships}: if the objects have proper relationships to each other (\eg{}, for a lunar eclipse, the earth should be between the sun and the moon).
We have an expert rate both stages on a Likert scale from 1 to 5 for 25 layouts/images.

\noindent \textbf{Pairwise preference: Image-text alignment \& object relationships.}
We conduct a human analysis comparing our \method{} framework to \sdonefour{} on 50 diagrams.
We choose \sdonefour{} fine-tuned on \dataset{}
for comparison because it is the closest baseline to our \method{} and also shows the strongest results among the baselines (see \cref{subsec:main_results}).
We ask crowd-sourced annotators (20 unique annotators) from Amazon Mechanical Turk\footnote{\url{https://www.mturk.com}} to evaluate the diagrams generated for each prompt (see appendix for setup details).

\section{Results and Discussion}
\label{sec:results}

We show our primary
quantitative results (\cref{subsec:main_results}),
human evaluation on pairwise preference study and error analysis (\cref{subsec:human_eval_results}),
qualitative analysis (\cref{subsec:qualitative_results}),
additional analysis about open-domain generation, multi-platform vector graphic diagram generation,
human-in-the-loop \plan{} editing,
(\cref{subsec:additional_analysis_results}),
and an ablation of different ablations for \plan{} generation (\cref{subsec:different_llms}).
See appendix for more analysis on \generator{} setups and multimodal planner/auditor LLMs.

\subsection{Quantitative Results}
\label{subsec:main_results}

\noindent\textbf{VPEval.}
\Cref{tab:main_results} left block shows the VPEval results.
For both \sdonefour{} and \vpgen{} baselines,
fine-tuning improves the score for object skill (\eg{}, 63.1 $\rightarrow$ 69.8 for \sdonefour{}, and 55.8 $\rightarrow$ 62.8 for \vpgen{}),
but for \vpgen{} count, it decreases scores (32.9 $\rightarrow$ 27.8).
For relationships, both models improve (79.3 $\rightarrow$ 81.9 for \sdonefour{}, and 72.8 $\rightarrow$ 76.3 for \vpgen{}).
For text, both models achieve 0 scores before and after fine-tuning.
Our \method{} outperforms both zeroshot and fine-tuned baselines on both overall and skill-specific VPEval scores,
showcasing the strong layout control, object relationship representation, and accurate text rendering capability of our diagram generation framework.

\begin{table}[t]
  \centering
    \resizebox{\textwidth}{!}{
  \begin{tabular}{l ccccc cc cc}
    \toprule
    \multirow{2}{*}{Methods} & \multicolumn{5}{c}{VPEval (\%) $\uparrow$} & \multicolumn{2}{c}{Captioning $\uparrow$} & \multicolumn{2}{c}{CLIPScore $\uparrow$} \\
    \cmidrule(lr){2-6} \cmidrule(lr){7-8} \cmidrule(lr){9-10}
     & Object & Count & Text & Relationships & Overall & CIDEr & BERTScore & Img-Txt & Img-Img \\
    \midrule
    \textit{Zeroshot} \\
    \sdonefour{} & 63.1 & 27.8 & 0.0 & 79.3 & 39.0 & 7.7 & 87.5 & 27.3 & 65.3 \\
    \vpgen{} & 55.8 & 32.9 & 0.0 & 72.8 & 39.3 & 6.1 & 87.2 & 25.6 & 61.7 \\
    \autotikz{} & 24.9 & 34.2 & 5.5 & 67.7 & 31.0 & 12.2 & 86.9 & 24.7 & 64.5 \\
    \midrule
    \textit{Fine-tuned} \\
    \sdonefour{} & 69.8 & 35.4 & 0.0 & 81.9 & 45.1 & 18.2 & 88.5 & 30.1 & 68.1 \\
    \vpgen{} & 62.8 & 27.8 & 0.0 & 76.3 & 39.7 & 4.2 & 86.9 & 26.4 & 61.9 \\
    \midrule
    \method{} (Ours) & \textbf{86.4} & \textbf{57.0} & \textbf{47.5} & \textbf{87.9} & \textbf{71.2} & \textbf{26.4} & \textbf{89.4} & \textbf{32.1} & \textbf{73.9} \\
    \bottomrule
  \end{tabular}
  }
  \caption{Comparison of \method{} to existing text-to-image generation baseline models. On all metrics, \method{} outperforms the baselines, indicating that our method is more effective for generating accurate diagrams.
  }
  \label{tab:main_results}
\end{table}

\noindent\textbf{Captioning and CLIPScore.}
\Cref{tab:main_results} middle block shows captioning scores (with LLaVA 1.5), and the right block shows CLIPScore (with CLIP-ViT L/14).
Our \method{} outperforms both the zeroshot and fine-tuned baselines, indicating our generated diagrams have more relevant information to the input prompt (which is a critical aspect of diagrams), have better image-text alignment, and more closely resemble the ground-truth diagrams (image-image) than the baselines.
Our \method{} significantly outperforms both fine-tuned \vpgen{} (26.4 \vs{} 4.2) and fine-tuned \sdonefour{} (26.4 \vs{} 18.2) for CIDEr and also achieves a few higher points on BERTScore.
For both CLIPScore's, \method{} has improvement over fine-tuned \sdonefour{} (32.1 \vs{} 30.1 and 73.9 \vs{} 68.1).

\subsection{Human Evaluation}
\label{subsec:human_eval_results}

\begin{table}[t]
\centering
\resizebox{\linewidth}{!}{
    \begin{tabular}{cc cc cc}
        \toprule
        \multicolumn{4}{c}{Stage 1: Diagram Planning (with GPT-4)} & \multicolumn{2}{c}{Stage 2: Diagram Generation (with \generator{})}\\
        \cmidrule(lr){1-4} \cmidrule(lr){5-6} 
        \multicolumn{2}{c}{Initial \plan{}} & \multicolumn{2}{c}{Diagram plan after refinement} & \multicolumn{2}{c}{Final Diagram} \\
        \cmidrule(lr){1-2} \cmidrule(lr){3-4} \cmidrule(lr){5-6} 
        Objects Presence ($\uparrow$) & Object Relations ($\uparrow$) & Objects Presence ($\uparrow$) & Object Relations ($\uparrow$) & Objects Presence ($\uparrow$) & Object Relations ($\uparrow$) \\
        \midrule
        4.96 & 4.56 & 4.96 & 4.72 & 2.96 & 3.36 \\
        \bottomrule
    \end{tabular}
}
\caption{Step-wise error analysis of \method{} on 25 \dataset{} test prompts.
We use a Likert scale (1-5) to evaluate object presence and object relations of the \plan{} (before and after refinement) and the final generated diagram.
}
\label{table:error_analysis}
\end{table}

\textbf{Step-by-step error analysis.}
\Cref{table:error_analysis}
shows that
our \plan{}s exhibit high scores on both object presence (4.96) and object relationship (4.56) even before the refinement, and refinement increases the object relationship scores even further (4.56 $\rightarrow$ 4.72) by adjusting the entity layouts from the initial \plan{}
(see
appendix
for refinement examples).
During the diagram generation stage, the object presence and relationship scores decrease.
This indicates that our \method{} could generate more accurate diagrams, once we have access to a layout-guided image generation backbone model stronger than \sdonefour{} architecture.

\begin{wraptable}[8]{r}{7.8cm}
  \centering
  \vspace{-10pt}
  \resizebox{\linewidth}{!}{%
  \begin{tabular}{lccc}
    \toprule
    \multirow{2}{*}{Evaluation category} & \multicolumn{3}{c}{Human Preference (\%) $\uparrow$} \\
    \cmidrule(lr){2-4}
    
     & \method{} & SD v1.4 & Tie \\
    \midrule
    Image-Text Alignment & \textbf{68} & 24 & 8 \\
    Object Relationships & \textbf{62} & 34 & 4  \\
    \bottomrule
  \end{tabular}
  }
\caption{
Human preference study on generated diagrams: \method{} \vs{} SD v1.4.
}
\label{tab:human_evaluation}
\end{wraptable}

\textbf{Human preference.}
We conduct a human preference study, comparing our \method{} and fine-tuned \sdonefour{} (SD v1.4) in image-text alignment and object relationships.
As shown in \Cref{tab:human_evaluation},
our \method{} achieves a higher preference than \sdonefour{} in both image-text alignment (68\% vs 24\%) and object relationships (62\% vs 34\%) criteria.

{
\begin{figure}[t]
    \centering
    \includegraphics[width=\textwidth]{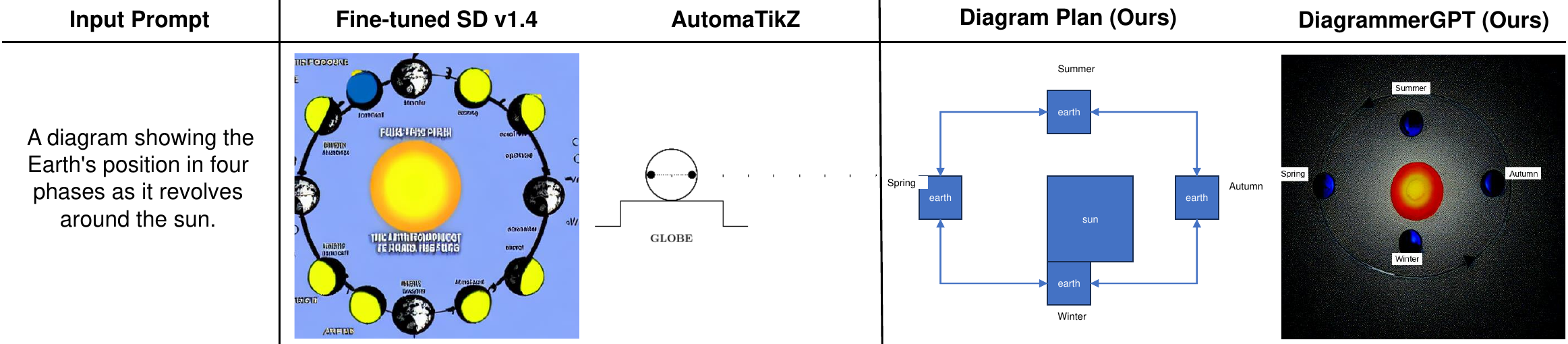}
    \caption{
    Example diagram generation results from baselines (fine-tuned \sdonefour{} and \autotikz{}) and our \method{} on the \dataset{} test split. Our \method{} correctly follows the caption while the baselines make several errors.
    }
    \label{fig:qualitative_examples_half}
\end{figure}
}

\subsection{Qualitative Analysis}
\label{subsec:qualitative_results}

\noindent \textbf{Comparison with baselines.}
\cref{fig:qualitative_examples_half} shows example diagrams generated by the baselines (\sdonefour{} and \autotikz{}) and our \method{} (both \plan{} and final generation diagram) on the \dataset{} test split (see appendix for additional example).
Our diagram plan strongly reflects the prompt, and the final diagram is more aligned with the input prompt than the baselines.
In the \cref{fig:qualitative_examples_half} example, our diagram correctly shows the earth in four phases revolving around the sun.
\sdonefour{} either over-generates objects in the image (\eg{}, too many earths), and \autotikz{} fails to generate a proper diagram.
Although our generated \plan{}s are generally correct, sometimes \generator{} can fail to properly follow all aspects. As noted in \cref{subsec:human_eval_results}, our \generator{} can improve once a better backbone becomes available.

\noindent\textbf{Diagram plan refinement.}
We analyze how our diagram refinement step (see \cref{subsec:stage1_method}) improves the diagram plans.
The refinement can fix minor mistakes like missing connections in a circuit or objects being placed in the wrong location.
We show examples in the appendix.

\subsection{Additional Analysis: Open-Domain, Open-Platform, and Human-in-the-Loop}
\label{subsec:additional_analysis_results}

\noindent \textbf{Open-domain diagram generation.}
We demonstrate that our planner LLM can extend its capabilities to unseen domains beyond the three areas (astronomy/biology/engineering) given in the in-context examples (30 total examples, 10 from each of astronomy/biology/engineering).
As shown in \cref{fig:open_domain_generation_half},
from an input prompt in the earth science domain,
our planner LLM generates fairly accurate layouts, and our \generator{} can generate a diagram following the layouts.
We also compare the recently released DALL-E 3~\citep{OpenAI2023DALLE3} model and find that it generally produces images with good aesthetic style but tends to generate diagrams with redundant objects (\eg{}, excessive text descriptions or objects). It also struggles with creating diagrams that adhere to a prompt (\eg{}, generating incorrect layers in the earth example). The DALL-E 3 system card~\citep{OpenAI2023DALLE3} also notes that DALL-E 3 tends to generate inaccurate information in diagrams.
See appendix for more examples.

{
\begin{figure}[t]
    \centering
    \includegraphics[width=\textwidth]{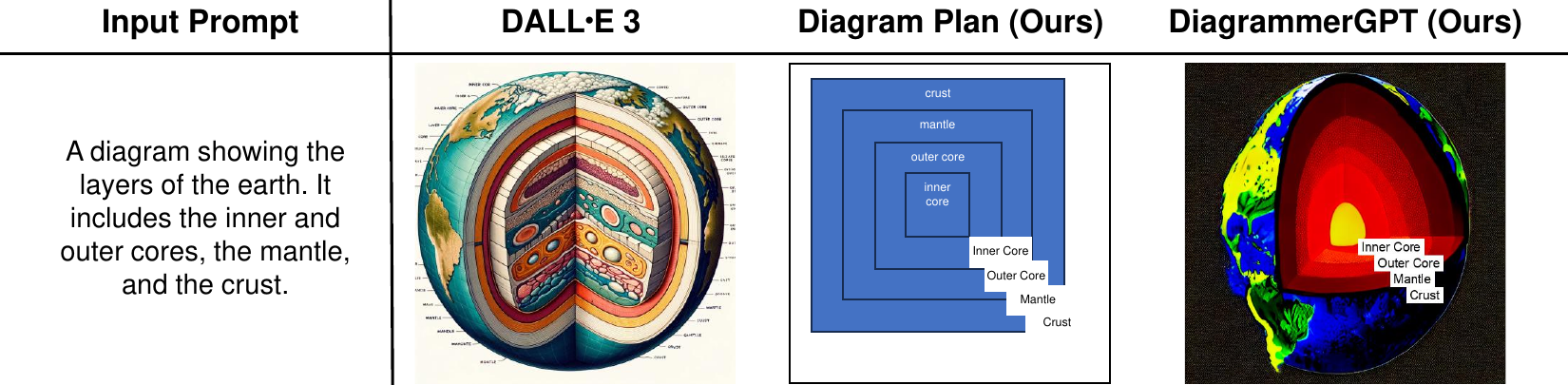}
    \caption{
    Open-domain diagram generation examples with DALL-E 3 and \method{}.
    }
    \label{fig:open_domain_generation_half}
\end{figure}
}

{
\begin{figure}[t]
    \centering
    \includegraphics[width=\textwidth]{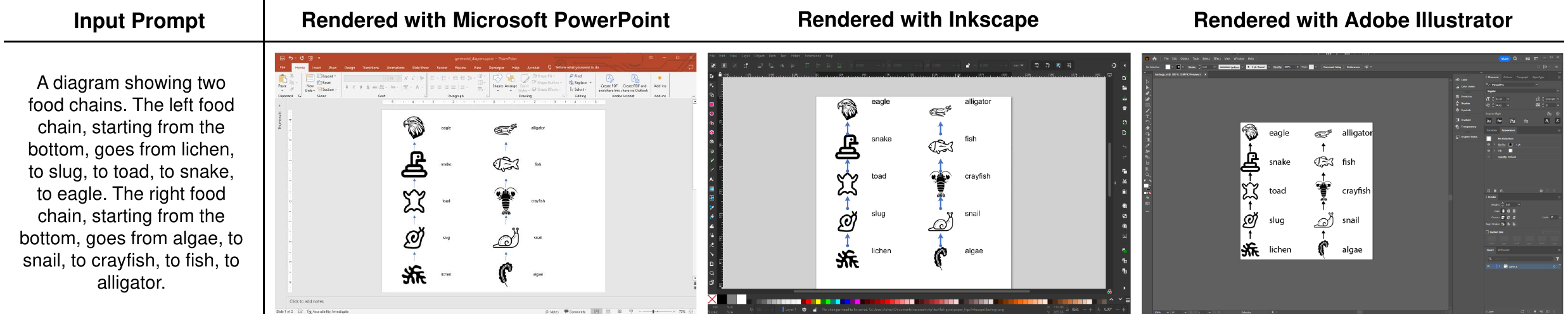}
    \caption{
    Examples of vector graphic diagrams generated from \plan{}s.
    }
    \label{fig:vector_diagram_examples_half}
\end{figure}
}

\noindent\textbf{Vector graphic diagram generation in different platforms.}
Although our primary focus is on a pixel-level diagram generation pipeline with \generator{}, our \plan{}s can also
be used for
vector graphic diagrams,
with multiple platforms like 
Microsoft PowerPoint, Inkscape,
and Adobe Illustrator.
\cref{fig:vector_diagram_examples_half} presents an example of vector graphic diagrams.
The example delivers promising results by effectively conveying the crucial information and layouts described in the input text prompts (see additional example in the appendix).
However, it also exhibits certain limitations:
(1) inconsistency in icon styles;
(2) limited icon retrieval capability.
As the diagrams are editable via these same platforms, these limitations can be addressed by end-users editing the diagrams to their liking (see below).

{
\begin{figure}[h]
    \centering
    \includegraphics[width=\textwidth]{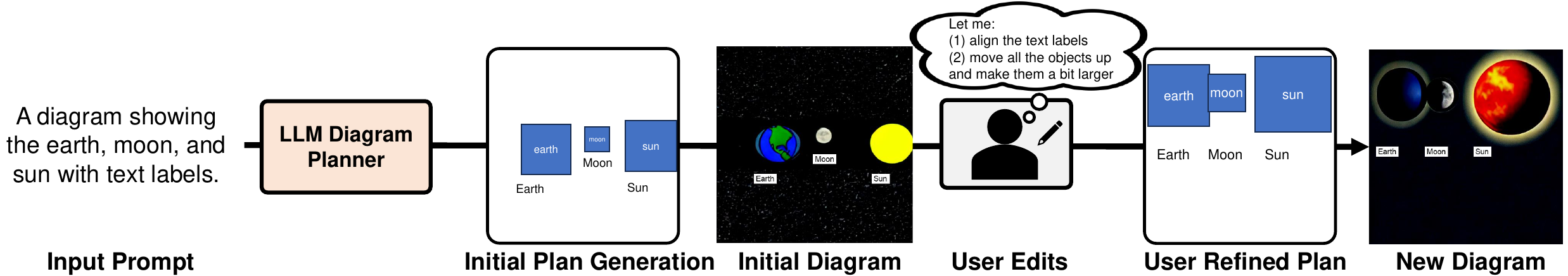}
    \caption{
    Illustration of human-in-the-loop \plan{} editing. 
    \method{} first provides an initial \plan{} with the corresponding generated diagram, users can then review the generated layouts/diagrams and make adjustments based on their needs.
    }
    \label{fig:human_in_the_loop}
\end{figure}
}

\noindent\textbf{Human-in-the-loop \plan{} editing.}
With the \plan{}s being rendered in vector graphic platforms, as mentioned above, our \method{} can provide an editable \plan{}, allowing for human-in-the-loop editing.
As illustrated in \cref{fig:human_in_the_loop}, \method{} first generates an initial \plan{} along with the rendered image.
Users can then review the generated layouts/diagrams and make adjustments based on their needs/wants (\eg{}, move the objects, add/remove objects, adjust object sizes, etc.).
With the human-refined diagram plan, users can either keep it in vector format and use icons (as mentioned in the previous paragraph) or give it back to \generator{} and then create pixel-level diagrams, resulting in diagrams/layouts that are better suited to end-users requirements.

\begin{wraptable}[11]{r}{7.8cm}
\vspace{-10pt}
  \centering
    \resizebox{0.5\columnwidth}{!}{
  \begin{tabular}{lccccccccccc|c}
    \toprule

    \multirow{2}{*}{LLM} & \multicolumn{3}{c}{Layout Recall} \\
    \cmidrule(lr){2-4}
    
     & Object & Text & Overall \\
    \midrule
    LLaMA3 8B & 24.9 & 32.2 & 29.2 \\
    \vpgen{} (Vicuna 13B) & 30.6 & 0 & 12.9 \\
    GPT-3.5 Turbo & 82.5 & 54.6 & 74.7 \\
    GPT-4 (default) & \textbf{84.1} & \textbf{60.1} & \textbf{78.4} \\
    \bottomrule
  \end{tabular}
  }
  \caption{
  Ablation of different LLMs for the planner and \auditor{}.
  }
  \label{tab:llm_ablation_results}
\end{wraptable}

\subsection{Different LLMs for \plan{} generation}
\label{subsec:different_llms}
We also experiment with different LLMs such as GPT-3.5 Turbo~\citep{chatgpt}, GPT-4~\citep{OpenAI2023GPT4TR}, LLaMA 3 (8B)~\citep{llama3modelcard}, and LLaMA2-Chat (13B)~\citep{Touvron2023Llama2}.
We compare them with `layout recall' -- how often each ground-truth object/text label can be found in the predicted \plan{}.
We first obtain bipartite matching of objects/text labels between prediction and ground-truth via Hungarian matching algorithm, using BLEU-1~\citep{Papineni2002BleuAM} as the matching score.
Then we calculate layout recall metric: $\frac{\sum_{n=1}^Nmatch(obj_n)}{N}$, where $N$ is the total number of ground-truth object/text labels, $obj_n$ is nth the ground-truth object/text label, and $match$ outputs 1 if the object/text label is matched in the predicted \plan{} and 0 if not.

\Cref{tab:llm_ablation_results} shows a comparison of four LLMs: LLaMA3 8B, GPT-3.5 Turbo, GPT-4, and Vicuna 13B (\vpgen{} checkpoint).
As indicated by GPT-3.5 Turbo's and GPT-4's high `overall' performance (74.7 and 78.4 respectively), both models are capable of generating accurate diagrams; however, as GPT-4 is slightly better in both the `object' and `text' metrics, we use GPT-4 as our main LLM. LLaMA3 and fine-tuned Vicuna are not able to perform well for the task with both getting an overall score below 30.

\section{Conclusion}
\label{sec:conclusion}

In this work, we propose \method{}, a novel two-stage text-to-diagram generation framework that leverages the knowledge of LLMs for planning and refining the overall \plan{}s.
We demonstrate that our \method{}
achieves more semantically accurate layouts in diagram generation than baseline models in both quantitative and qualitative analysis.
In addition, we provide comprehensive human error analysis, ablation studies, and analysis about
open-domain diagram generation,
vector graphic diagram generation,
human-in-the-loop \plan{} editing, and
multimodal planner/\auditor{} LLMs.
We hope our work can inspire future research on diagram generation.

\section*{Acknowledgments}
We thank the reviewers for the thoughtful discussion and feedback. This work was supported by DARPA ECOLE Program No. HR00112390060, NSF-AI Engage Institute DRL-2112635, DARPA Machine Commonsense (MCS) Grant N66001-19-2-4031, ARO Award W911NF2110220, ONR Grant N00014-23-1-2356, Accelerate
Foundation Models Research program, and a Bloomberg Data Science Ph.D. Fellowship. The views contained in this article are those
of the authors and not of the funding agency.

{
    \small
    \bibliographystyle{colm2024_conference}
    \bibliography{main}
}

\appendix

\renewcommand\thesection{\Alph{section}}

\section*{Appendix}

In this appendix, we provide
LLM prompt templates and \plan{}s used in the diagram planning stage (\cref{appendix:llm_input_prompts}),
\dataset{} collection details,(\cref{appendix:dataset_collection_details}),
additional \generator{} details (\cref{appendix:gen_training_details}),
experimental setup details (\cref{appendix:experimental_setups}),
human evaluation setup details (\cref{appendix:human_evaluation_details}),
additional results/analysis/visualizations/ablations details (\cref{appendix:additional_results}),
and limitations (\cref{appendix:limitations}).

\section{LLM Prompt Templates and Diagram Plans}
\label{appendix:llm_input_prompts}

\paragraph{Prompt templates.}

In \cref{fig:planning_prompt} and \cref{fig:corrector_prompt},
we show the prompt templates for the planner LLM and \auditor{} LLM, respectively.
For both LLMs we provide 10 in-context examples followed by the inputs.
For planner LLM, we give it the caption and the topic of the caption (\eg{}, astronomy, biology, \etc{}).
For \auditor{} LLM, we give the \plan{} generated by the planner and ask if there are any issues.

\paragraph{Diagram plans.}
A \textit{\plan{}} consists of three components:
(1) entities - a dense list of objects (\eg{}, larva in \cref{fig:diagram_plan}) and text labels (\eg{}, ``egg'' in \cref{fig:diagram_plan});
(2) relationships - complex relationships between entities (\eg{},
object-object relationship ``[obj\_0] \textit{has an arrow to} [obj\_1]''
or
object-text label relationship ``[text\_label\_0] \textit{labels} [obj\_0]'');
(3) layouts - 2D bounding boxes of the entities (\eg{}, ``[obj\_0]: $[20, 30, 14, 14]$'' in \cref{fig:diagram_plan}).
For object-object relationships, we utilize two types: line and arrow (a line with explicit start and end entities), which are useful when specifying object relationships in diagrams such as flow charts or life cycles.
For object-text label relationships, we specify which object each label refers to.
For layouts, we use the $[x, y, w, h]$ format for 2D bounding boxes, whose coordinates are normalized and integer-quantized within $\{0, 1, \cdots 100\}$, in accordance with VPGen~\citep{Cho2023VPT2I}.

\paragraph{Full \plan{} example.}
\cref{fig:full_diagram_plan_example} shows 
an example of a fully generated \plan{} by our LLM planner for the prompt ``A diagram showing the life cycle of a butterfly, going from an egg to larva to pupa to an adult butterfly and repeating.''
This \plan{} corresponds to the second example in \cref{fig:qualitative_examples}.

{
\small
\begin{figure}
    \centering
    \includegraphics[width=0.99\textwidth]{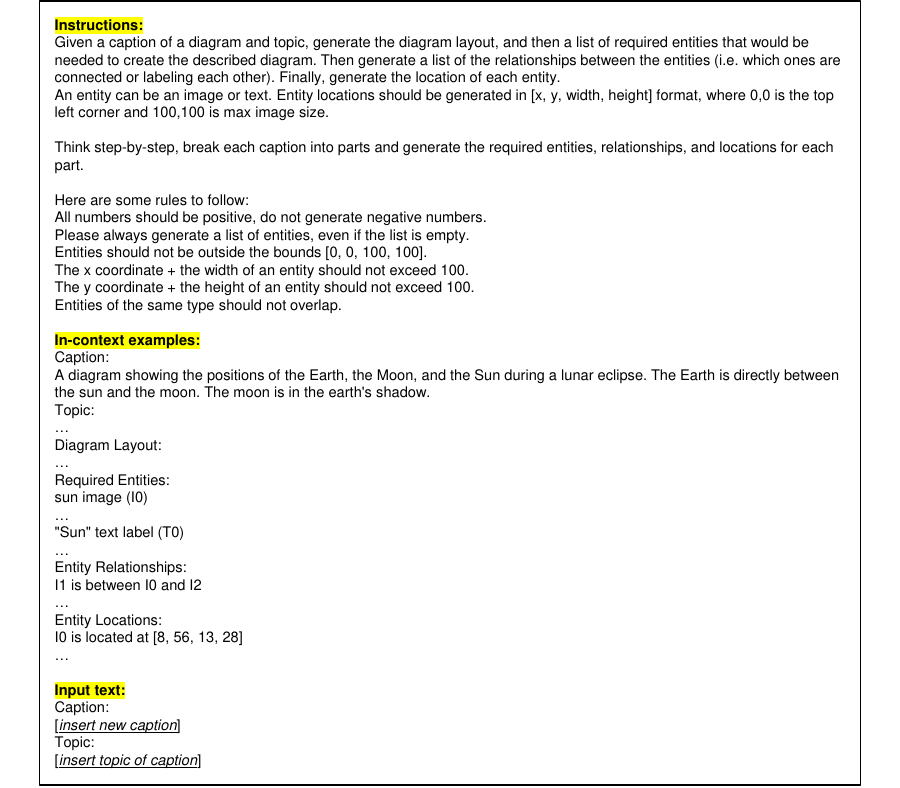}
    \caption{Input prompt given to the planner LLM during the initial \plan{} generation step.}
    \label{fig:planning_prompt}
\end{figure}
}

{
\small
\begin{figure}
    \centering
    \includegraphics[width=0.99\textwidth]{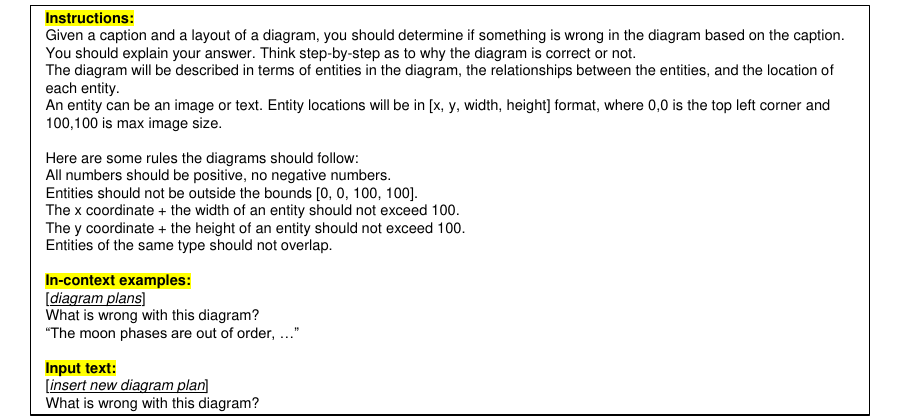}
    \caption{Input prompt given to the \auditor{} LLM during the \plan{} refinement step.}
    \label{fig:corrector_prompt}
\end{figure}
}

\begin{figure}[t]
\centering
\framebox{
\begin{minipage}{.8\linewidth}
\footnotesize
\bfit{Required Entities:}\\
egg image (I0)\\
larva image (I1)\\
pupa image (I2)\\
adult butterfly image (I3)\\
``egg'' text label (T0)\\
``larva'' text label (T1)\\
``pupa'' text label (T2)\\
``adult butterfly'' text label (T3)\\
\bfit{Entity Relationships:}\\
I3 has an arrow to I0\\
T3 labels I3\\
I1 has an arrow to I2\\
T0 labels I0\\
I0 has an arrow to I1\\
T2 labels I2\\
I2 has an arrow to I3\\
T1 labels I1\\
\bfit{Entity Locations:}\\
I0 is located at [24, 50, 14, 14]\\
I1 is located at [50, 74, 14, 14]\\
I2 is located at [74, 50, 14, 14]\\
I3 is located at [50, 24, 14, 14]\\
T0 is located at [20, 44, 10, 4]\\
T1 is located at [44, 80, 10, 4]\\
T2 is located at [80, 44, 10, 4]\\
T3 is located at [44, 20, 10, 4]
\end{minipage}
}
\caption{Example \plan{} from our LLM planner for the prompt ``A diagram showing the life cycle of a butterfly, going from an egg to larva to pupa to an adult butterfly and
repeating.''}
\label{fig:full_diagram_plan_example}
\end{figure}

\paragraph{LLM API Costs}
\label{appendix:api_costs}
The average input token length for the planner stands at 4.6K, while the average output token length is 0.5K. Generating a \plan{} using GPT-4 costs \$0.17 USD.

\section{\dataset{} Collection Details}
\label{appendix:dataset_collection_details}

To create the \dataset{} dataset described in main paper \cref{subsec:dataset_collection}, we employ LLaVA 1.5~\citep{liu2023improvedLLava}, a state-of-the-art multimodal language model, to generate captions and bounding box region descriptions in AI2D diagrams.
The original AI2D dataset provides annotations for diagrams, including titles, bounding boxes for object/text labels, and object linkages (\eg{} arrows/lines between objects).
However, since the dataset is designed for the diagram question-answering task rather than diagram generation, the diagram titles are often too short and don't provide enough information to produce meaningful \plan{}s.
Additionally, the dataset doesn't include descriptions for each object (\ie{}, each object bounding box is labeled simply as `blob').
To generate captions for each AI2D diagram, we present the diagram to LLaVA 1.5 and prompt it with the question ``What is this diagram showing?''.
To collect region descriptions of the bounding boxes in AI2D, we first overlay the bounding box annotations of each object on the diagram, by assigning each box a label (\ie{}, box 1 would get label ``B1'', box 2 would get label ``B2'', \etc{}.).
Then we provide this annotated image to LLaVA 1.5 and ask the model to describe each box's content (\eg{}, ``what is the object labeled by `B1'"?).
An example of this annotation is shown in~\cref{fig:diagram_dataset}.

To ensure the quality of \dataset{} annotations, we perform a human evaluation of 50 LLaVA 1.5 annotations.
For captioning, LLaVA generates very good captions 80\% of the time, and for labeling, LLaVA generates very good bounding box region descriptions 68\% of the time.
We find that when LLaVA makes a captioning error, it is usually minor points, and for bounding box region errors, sometimes it may give nearby boxes the same description.
While automatic annotations have some errors, we find that they are good enough for the domain adaption of \generator{};
However, for the test set of \dataset{} and in-context learning annotations for the planner LLM, we manually annotate them to ensure correctness (\cref{subsec:dataset_collection}).

\section{Additional \generator{} Details}
\label{appendix:gen_training_details}

\paragraph{Training.}
The parameters of \generator{} are initialized from the GLIGEN (Box+Text checkpoint)\footnote{\url{https://github.com/gligen/GLIGEN}}, and trained on the \dataset{} dataset (see main paper \cref{subsec:dataset_collection} and \cref{appendix:dataset_collection_details} for details) for 15k steps (batch size of 5 per GPU), which takes 12 hours with 8 A6000 GPUs (each 48GB memory). Other hyperparameters include:
\begin{itemize}
    \item Optimizer: AdamW~\citep{Loshchilov2017FixingWD}
    \item Learning Rate: 5e-5
    \item Warmup Steps: 2500
    \item Image Size: 512x512
\end{itemize}

\paragraph{GLIGEN gated self-attention layers.}
The high-level intuition of the gated self-attention layer is similar to cross-attention layers,
which are used to incorporate extra information (\eg{}, text, layouts) into the model.
The cross-attention layer in the image generation backbone is used to incorporate text tokens for text-guided image generation, while the gated self-attention layer takes the grounding tokens for layout-guided image generation.
Additional details are clarified in the GLIGEN paper~\citep{li2023gligen}.

\section{Experimental Setups}
\label{appendix:experimental_setups}

\subsection{Metrics}

\paragraph{VPEval (Objects, Counts, Relationships, Texts).}
We evaluate the diagrams in terms of the presence of objects (object evaluation), the number of objects (count evaluation), the correctness of spatial and connection relationships (relationship evaluation), and the presence of correct text labels (text evaluation) using the VPEval metric~\citep{Cho2023VPT2I}. VPEval works by first generating evaluation programs that call specific evaluation modules (\eg{}, Object, OCR, VQA) and then running the modules to evaluate the image. Each module evaluates different parts of the image (\eg{}, Object checks object presence).
For object, count, and text rendering evaluation, we use the ground-truth \plan{} to determine which evaluation programs to use.
For relationship evaluation, following VPEval, we use an LLM (GPT-4) to generate VQA questions that evaluate the spatial/connection relationships of objects in the diagrams (\eg{}, if the moon is in between the sun and earth or if a light bulb is connected to a battery).
While the original VPEval uses BLIP-2~\citep{Li2023BLIP2BL} as its visual reasoning model,
we employ
the recently released
LLaVA 1.5~\citep{liu2023improvedLLava} model as our visual reasoning model, as we find it
is more faithful in diagram question answering in our initial experiments and is an overall stronger model than BLIP-2~\citep{liu2023improvedLLava}.
In our initial experimentation, we found that LLaVA 1.5 sometimes struggles with relationship evaluation, so we fine-tune the model on relationships to ensure accurate evaluation.

For object evaluation, LLaVA determines if the object is present.
For count evaluation (\ie{}, if a diagram requires multiple instances of an object),
we ask LLaVA if there are exactly $N$ instances of the object in the diagram, where $N$ represents the count of that object in the ground-truth diagram.
For relationship evaluation, we ask LLaVA if the relation (spatial or connection) is true.
In our experiments, we find that LLaVA often can generate false positives during relation evaluation, as such we also show human evaluation for object relationships (main paper \cref{subsec:human_eval_setup}). 
For text evaluation, in alignment with VPEval, we utilize EasyOCR~\citep{EasyOCR2023} as the OCR model and check if the target text is detected.

\section{Human Evaluation Setup Details}
\label{appendix:human_evaluation_details}
We employ crowd-workers from Amazon Mechanical Turk (AMT)
for our human preference study. To ensure high-quality annotations, 
we set the following requirements for the workers:
they must possess an AMT Masters qualification, have completed more than 1000 HITs, maintain an approval rating above 95\%, and come from the United States, Great Britain, Australia, or Canada, given that our task is in English.
We pay workers \$0.06 to compare two diagrams (roughly \$14-15/hr).
For each prompt, we show diagrams generated by both our \method{} and the fine-tuned \sdonefour{}
(the order of diagrams are randomly shuffled every time to prevent selection biases)
and ask five annotators to indicate their preference based on
(1) the accuracy of the generated relationships between objects (\eg{}, spatial relationships and arrows/lines)
and (2) alignment to the input prompt (\eg{}, how well does the generated diagram reflect the input prompt).
Then, we take the agreement of the annotators.
The task is described to the annotators as such:
\begin{enumerate}
    \item Object Relationships is a measure of which diagram better captures the proper relationships of the objects (\ie{}, spacing/positioning of them, arrows/lines between them, \etc{}).
    \item Alignment is a measure of which diagram is a better representation of the input sentence.
\end{enumerate}
\cref{fig:human_eval_inferface} shows the interface provided to the annotators during the human evaluation process.

{
\begin{figure}
    \centering
    \includegraphics[width=.95\textwidth]{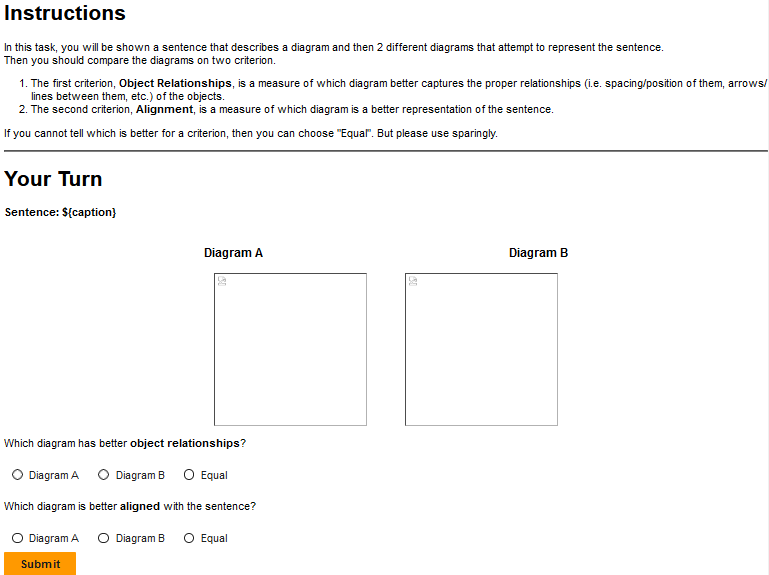}
    \caption{
    Interface provided to annotators for human evaluation.
    }
    \label{fig:human_eval_inferface}
\end{figure}
}

\section{Additional Experiment Results}
\label{appendix:additional_results}

{
\begin{figure}
    \centering
    \includegraphics[width=.95\textwidth]{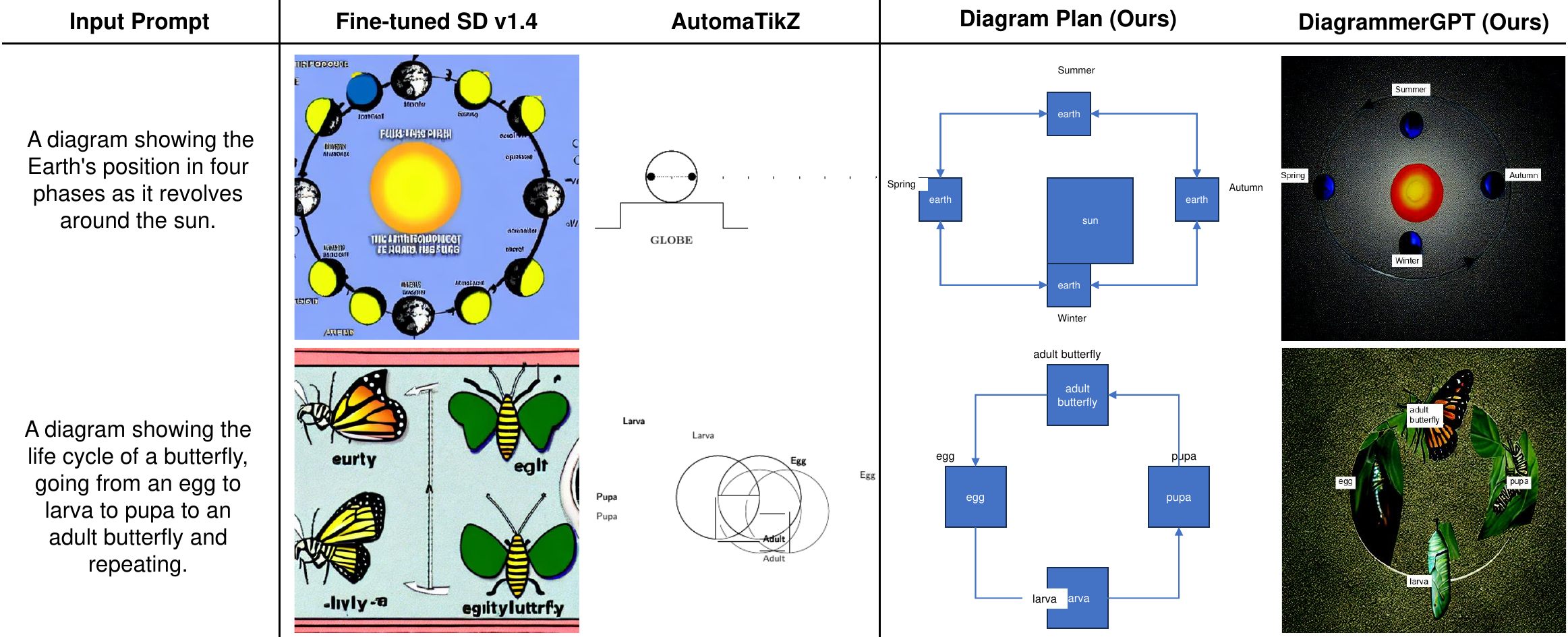}
    \caption{
    Example diagram generation results from baselines (fine-tuned \sdonefour{} and \autotikz{}) and our \method{} on the \dataset{} test split.
    In the first example, our \method{} correctly gets the object count right and has clear text, whereas \sdonefour{} overpopulates the entities orbiting around the sun. In the second example, our \method{} generates an accurate \plan{} and a diagram that mostly reflects the plan, whereas \sdonefour{} fails to show a life cycle (\ie{}, missing the egg, pupa, and larva). As noted in main paper \cref{subsec:human_eval_results}, once a better backbone becomes available, our \method{} can produce better diagrams based on the \plan{}s.
    \autotikz{} struggles to generate the proper layouts and objects for both examples.
    }
    \label{fig:qualitative_examples}
\end{figure}
}

{
\begin{figure}
    \centering
    \includegraphics[width=.9\textwidth]{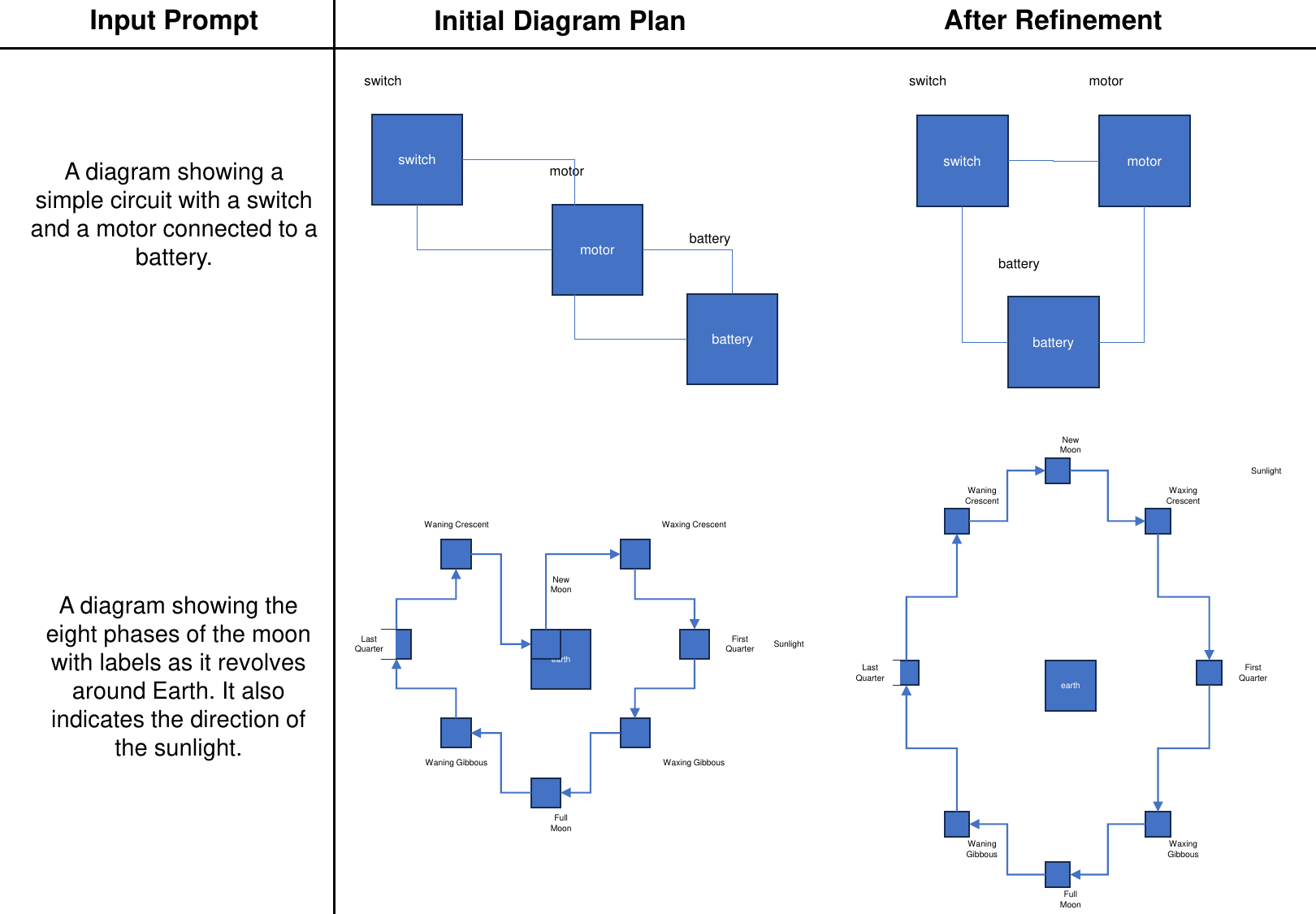}
    \caption{
    Examples from our diagram refinement step. Our \auditor{} LLM can help reorganize the connections between the components to be more clear in the first example and prevent overlaps of objects in the second example.
    }
    \label{fig:feedback_comparison}
\end{figure}
}

\subsection{Qualitative Results and Analysis}

\paragraph{Qualitative comparison to baselines.}
\cref{fig:qualitative_examples} shows example diagrams generated by the baselines (\sdonefour{} and \autotikz{}) and our \method{} (both \plan{} and final generation diagram) on the \dataset{} test split.
Our diagram plans strongly reflect the prompts and the final diagrams are more aligned to the input prompts.
In \cref{fig:qualitative_examples} top example, our diagram correctly shows the earth in four phases revolving around the sun and in the second example, our \plan{} correctly represents the life cycle of a butterfly and the generated diagram captures the circular flow of the \plan{} as well most aspects of the life cycle.
\sdonefour{} either over- or under-generates objects in the image (\eg{}, too many earths in the first example and missing egg/larva/pupa stages in the second example), and \autotikz{} fails to generate proper layouts and objects.
Although our generated diagram plans are generally correct, however, sometimes \generator{} can fail to properly follow all aspects (\eg{}, the egg is misdrawn and the larva/pupa are swapped in \cref{fig:qualitative_examples} bottom example). As noted in main paper \cref{subsec:human_eval_results}, once a better backbone becomes available, our \generator{} can produce better diagrams following the \plan{}s.

\paragraph{Diagram plan refinement.}
In \cref{fig:feedback_comparison}, we show how our diagram refinement step (see main paper \cref{subsec:stage1_method})
improves the diagram plans.
In the top example, the switch is not connected to the battery, thus does not affect the circuit.
After refinement, the connections are corrected so the switch is now also connected to the circuit and the layouts are adjusted to have a more straightforward flow.
In the bottom example, the moon phase of `New Moon' is too low and overlaps with the `Earth' object.
After refinement, there is no more overlap.

\subsection{Additional Analysis}

{
\begin{figure}[t]
    \centering
    \includegraphics[width=.9\textwidth]{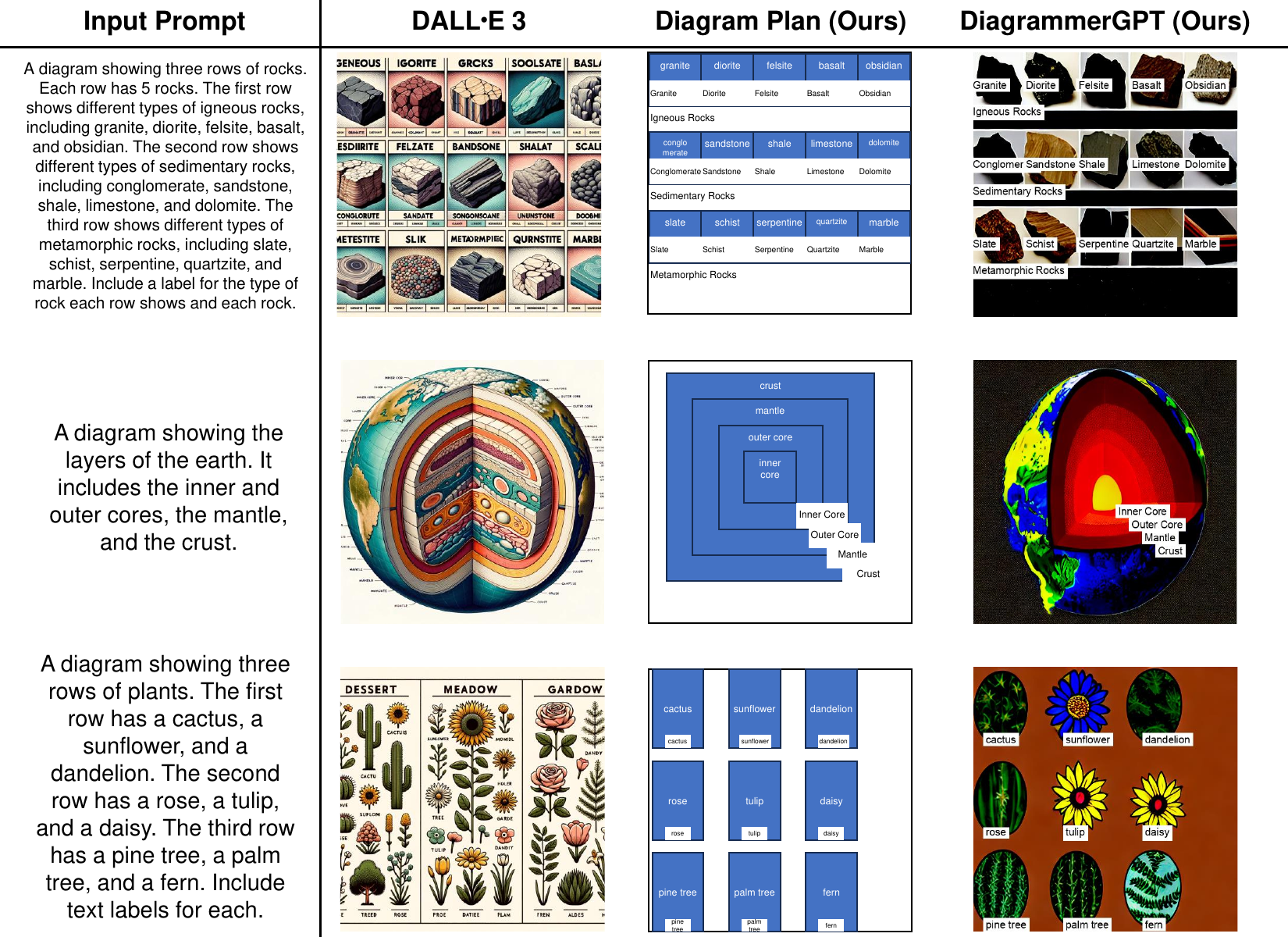}
    \caption{
    Examples of open-domain generation demonstrate that our \method{} can create diagrams that adhere to the input text prompts. Although DALL-E 3 yields images with superior visual quality, it tends to generate diagrams with redundant and crowded objects and also struggles to follow the prompt accurately (\eg{}, in the second example, it is not clear where the locations of layers such as the `inner core', `outer core', and `mantle' are. In the third example, it generates too many objects that are not in rows).
    }
    \label{fig:open_domain_generation_full}
\end{figure}
}

\paragraph{Open-domain diagram generation.} 
Our main diagram generation experiments are conducted on diverse domains such as astronomy, biology, and engineering which are included in the LLM planner's in-context examples.
However, given that the in-context examples do not encompass all diagram domains, we experiment with generating diagrams in areas not covered by our LLM in-context examples, such as geology and botany, to assess whether our \method{} maintains its ability to produce more accurate diagrams in previously unseen domains.

In \cref{fig:open_domain_generation_full}, we show examples of comparing our open-domain diagram generation to DALL-E 3. 
While our \generator{} struggles in some cases (like the third example), it is able to strongly adhere to the \plan{}.
\cref{fig:additional_examples} (bottom) also shows our LLM planner is easily able to generalize to completely new domains (\eg{}, neural networks and vacation planning).
As mentioned in main paper \cref{subsec:human_eval_results}, once a stronger layout-guided image generation model than GLIGEN with \sdonefour{} backbone is available, our \method{} can produce higher quality results.
We find that DALL-E 3 generally produces images with good aesthetic style but tends to generate diagrams with redundant and crowded objects (\eg{}, excessive unnecessary text descriptions in the rock and Earth examples, and an overabundance of plants in the third example). It also continues to struggle with creating accurate diagrams that adhere to a prompt (\eg{}, generating incorrect layers in the earth example and generating three columns of plants instead of three rows in the plant example). The DALL-E 3 system card~\citep{OpenAI2023DALLE3} also notes that DALL-E 3 tends to generate scientifically inaccurate information in diagrams.

\paragraph{Vector graphic diagram generation in different platforms.}
We render our \plan{}s
in Microsoft PowerPoint via VBA language,\footnote{\url{https://learn.microsoft.com/en-us/office/vba/api/overview/powerpoint}}
Inkscape\footnote{\url{https://inkscape.org}}
via a Python scripting extension\footnote{
\url{https://github.com/spakin/SimpInkScr}},
and Adobe Illustrator\footnote{\url{https://www.adobe.com/products/illustrator.html}} via JavaScript.
We represent objects using icons, which are retrieved via the Noun Project Icons API based on corresponding text descriptions.\footnote{\url{https://thenounproject.com/api/}}
\cref{fig:additional_examples} (top) and 
\cref{fig:vector_diagram_examples_full}
show additional examples of \plan{}s rendered in the other platforms.

{
\begin{figure}
    \centering
    \includegraphics[width=.8\columnwidth]{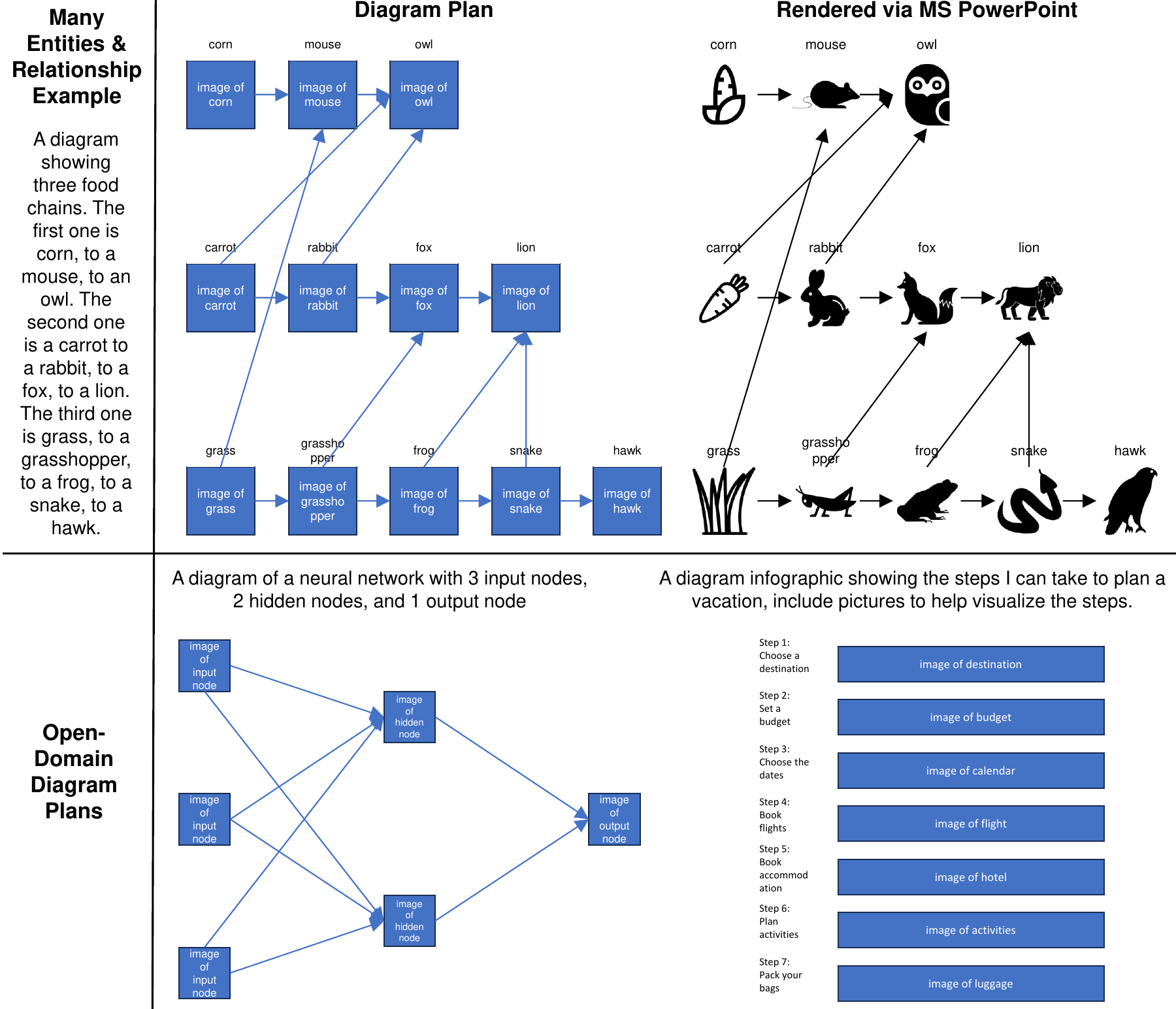}
    \caption{
    Additional examples. Our planner LLM is effective at generating dense (top) and open-domain (bottom) \plan{}s.
    }
    \label{fig:additional_examples}
\end{figure}
}

\begin{figure}[t]
    \centering
    \includegraphics[width=.9\columnwidth]{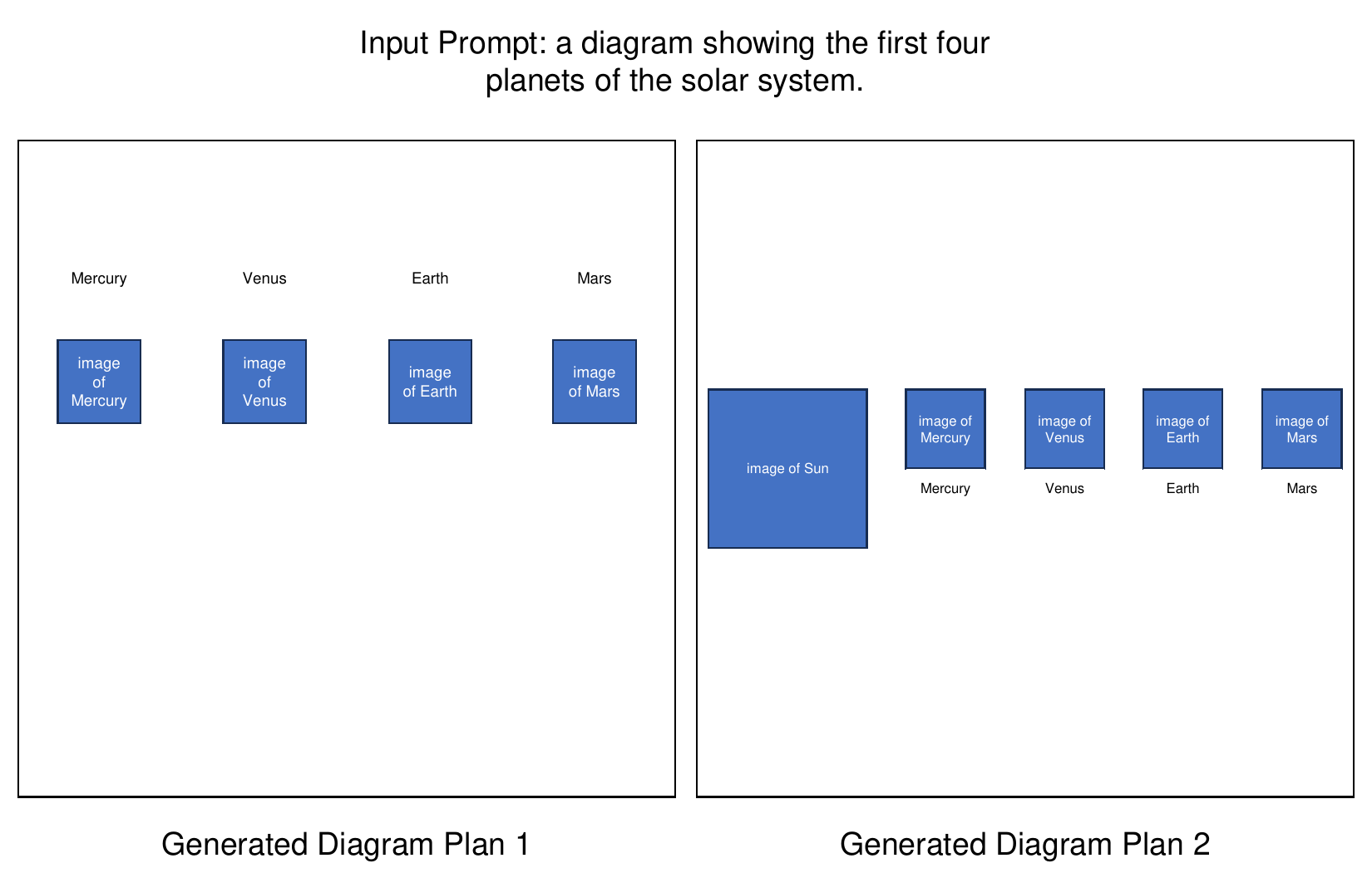}
    \caption{
    Example of two GPT-4 generated \plan{}s. Given the same prompt, GPT-4 can generate diverse plans between runs.}
    \label{fig:diverse_plans}
\end{figure}

{
\begin{figure}[t]
    \centering
    \includegraphics[width=0.99\textwidth]{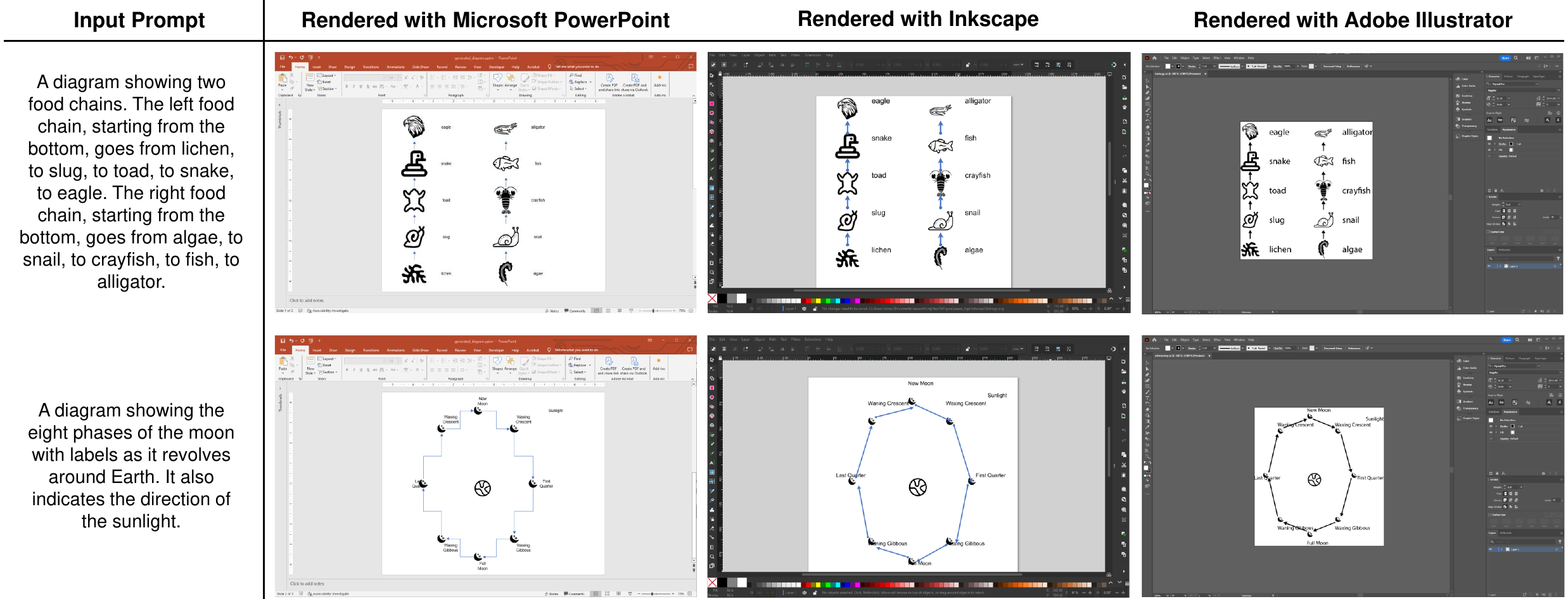}
    \caption{
    Examples of vector graphic diagrams generated with our \plan{}s and exported into Microsoft PowerPoint, Inkscape, and Adobe Illustrator.
    }
    \label{fig:vector_diagram_examples_full}
\end{figure}
}

\paragraph{GPT-4 \vs{} GPT-4Vision for \plan{} creation and refinement.}
As described in main paper, our \method{} employs a text-only GPT-4 model for diagram planning and refinement.
To explore whether a multimodal language model can offer improvements over text-only GPT-4,
we experiment with using the recently introduced GPT-4Vision (GPT-4V) model~\citep{OpenAI2023GPT4Vison} as the \textit{planner} and \textit{\auditor{}} LLM during the diagram generation and refinements steps.
As the GPT-4V model does not provide API access yet, we conduct a small-scale qualitative study via the ChatGPT web UI.
In our experiments, for the \plan{} creation stage (see main paper \cref{subsec:stage1_method}), GPT-4V does not provide improvements over text-only GPT-4.
In \cref{fig:gpt4v_comparison}, we present a comparison between the diagram plans generated by GPT-4 and GPT-4V.
GPT-4V does not produce \plan{}s that are better than text-only GPT-4, suggesting that our text-only representation is robust enough until better or fine-tuned versions of GPT-4V become available for diagrams. 
Similarly, during the diagram refinement step (see main paper \cref{subsec:stage1_method}),
we observed that GPT-4V tends to overestimate correctness when compared to text-only GPT-4,
further indicating the strength of our text representation.
\cref{fig:gpt4v_feedback_comparison} shows two examples comparing the models.
While text-only GPT-4 is not perfect, it can identify some errors, whereas GPT-4V says the diagram does not need improvement.

\begin{figure}[t]
    \centering
    \includegraphics[width=.7\columnwidth]{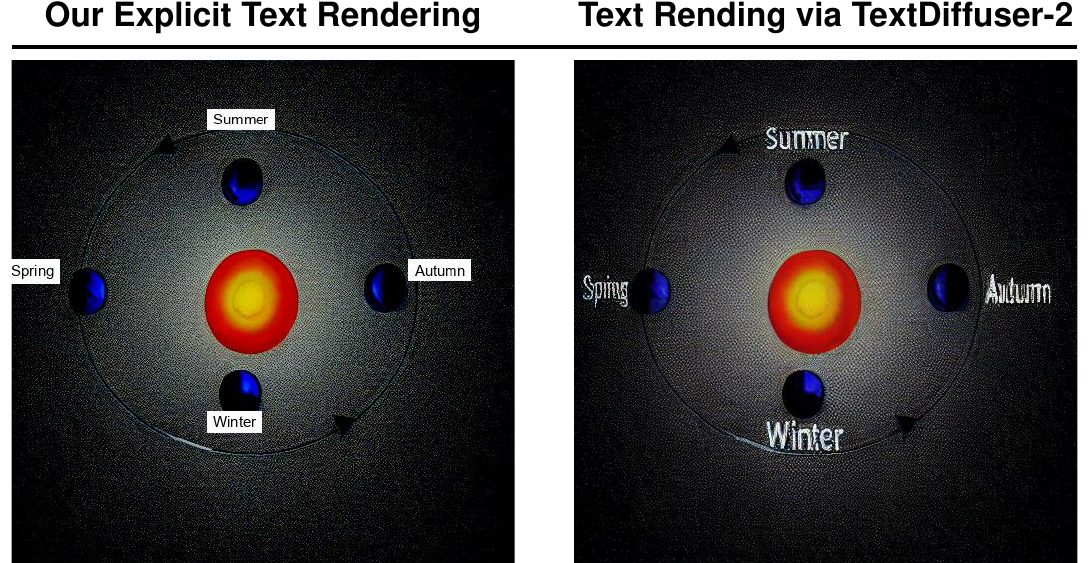}
    \caption{Explicit text rendering via \textpackage{} compared to model-based text rendering via TextDiffuser-2~\citep{chen2023textdiffuser2}.}
    \label{fig:text_rendering_examples}
\end{figure}

\paragraph{Using model based text rendering instead of \textpackage{}.}
We also experiment with using a model-based text renderer, TextDiffuser-2~\citep{chen2023textdiffuser2} instead of \textpackage{}.
\cref{fig:text_rendering_examples} shows that while TextDiffuser-2 is capable of producing good text labels, however, it can sometimes merge letters (\eg{}, the "mm" in summer).
\textpackage{} guarantees there is no rendering error (and can easily allow font color/size adjustments).
Due to the modular nature of \method{}, using a text rendering model can easily be incorporated if the end user wants.

{
\begin{figure}[t]
    \centering
    \includegraphics[width=0.77\textwidth]{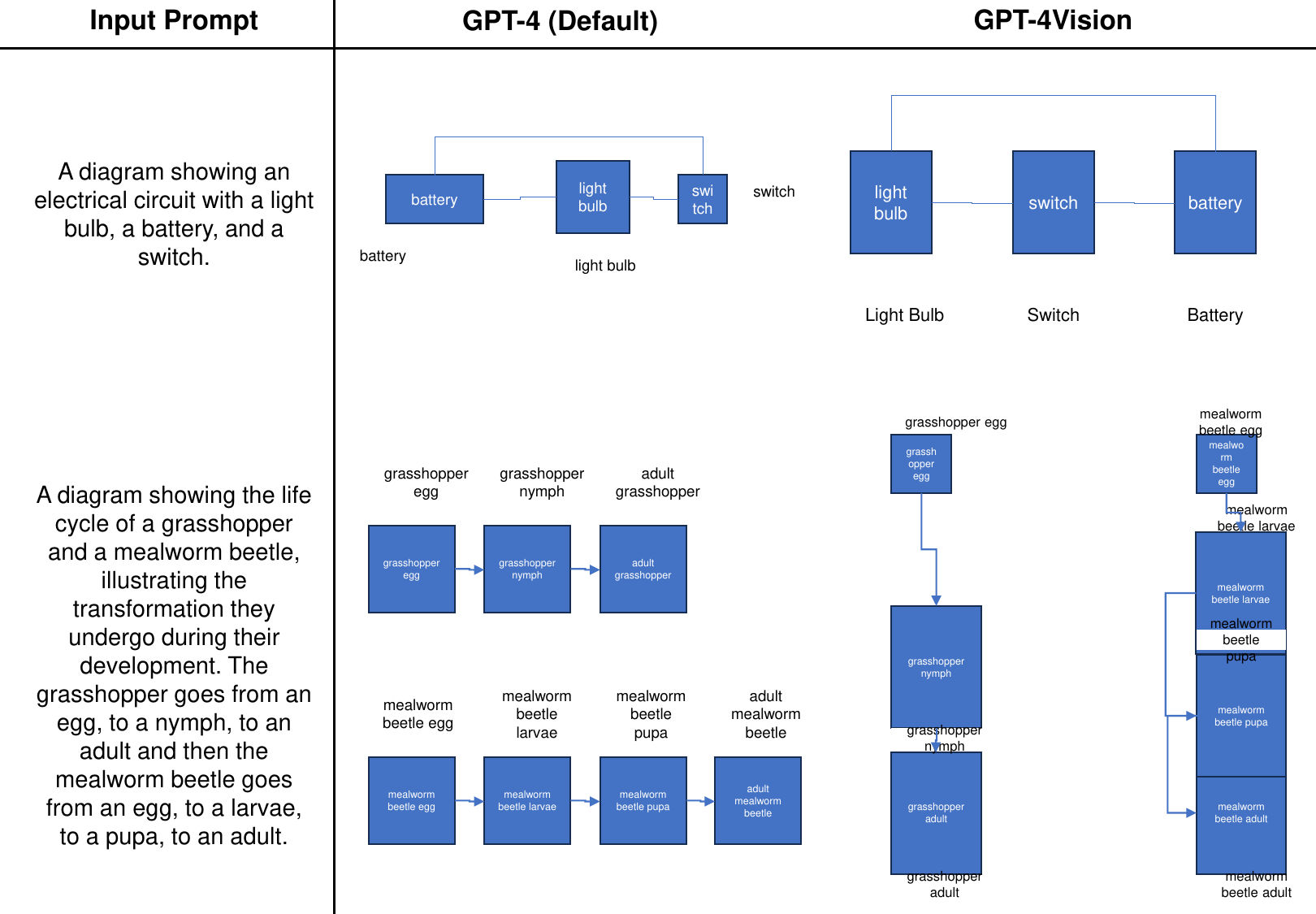}
    \caption{
    Comparison examples of \plan{}s generated by GPT-4 and GPT-4Vision (GPT-4V). GPT-4 creates \plan{}s that are sufficiently accurate in capturing the presence of objects and their relationships and GPT-4V does not provide plans that are better.
    }
    \label{fig:gpt4v_comparison}
\end{figure}
}

{
\begin{figure}[t]
    \centering
    \includegraphics[width=0.77\textwidth]{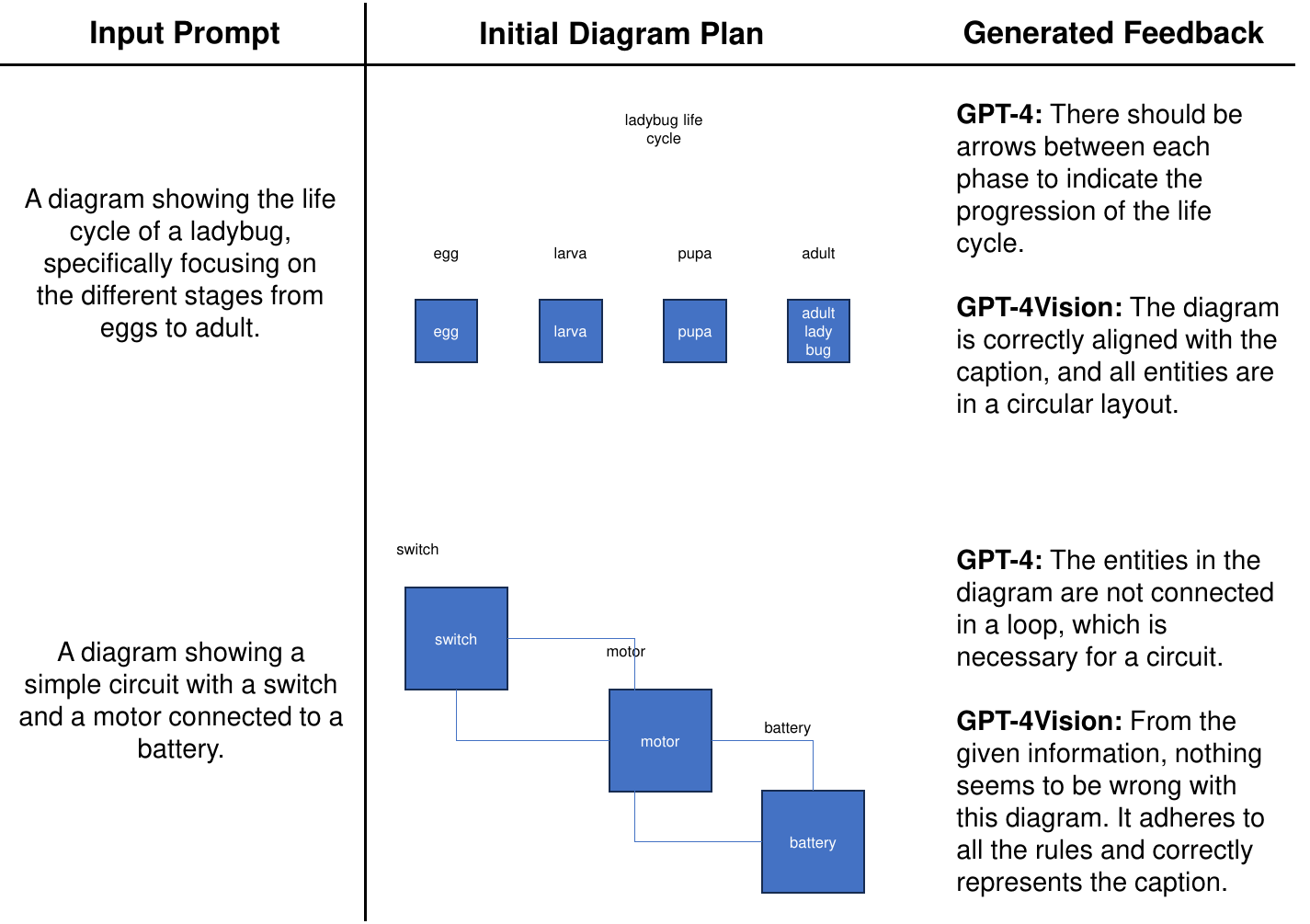}
    \caption{
    Comparison examples of GPT-4 \vs{} GPT-4Vision for \plan{} refinement. While text-only GPT-4 is not perfect, it can identify the errors, whereas GPT-4Vision says the diagram does not need improvement.
    }
    \label{fig:gpt4v_feedback_comparison}
\end{figure}
}

\paragraph{Is the LLM capable of generating diverse \plan{}s?}
We find that having an LLM generate the \plan{}s results in a wide diversity of diagrams. As we show in Fig. \ref{fig:qualitative_examples}, \ref{fig:feedback_comparison}, \ref{fig:open_domain_generation_full}, and \ref{fig:additional_examples}, the LLM can generate plans for many different prompts and domains. In \cref{fig:diverse_plans} we show an example of the LLM generating variations of the same prompt, further indicating the LLM (GPT-4 in our case) is capable of producing diverse \plan{}s.

\subsection{Ablation Studies}
\label{subsec:ablation_study_results}

We show ablation studies on our design choices with \generator{}: the number of denoising steps with layout guidance and whether to only update the gated self-attention layers parameters.

\begin{table}[h]
  \centering
    \resizebox{0.75\columnwidth}{!}{
  \begin{tabular}{lcccc}
    \toprule
    \multirow{2}{*}{\shortstack{\# Layout guidance steps}} & \multicolumn{4}{c}{VPEval} \\
    \cmidrule{2-5}
    & Object & Count & Text & Relationships \\
    \midrule
    $\alpha=5$ steps & \textbf{88.0} & 50.6 & \textbf{48.0} & 84.9 \\
    $\alpha=10$ steps & 86.4 & 49.4 & 47.0 & 86.6 \\
    $\alpha=15$ steps (default) & 86.4 & \textbf{57.0} & 47.5 & \textbf{87.9} \\
    \bottomrule
  \end{tabular}
  }
  \caption{Ablation of \# denoising steps with layout guidance. \generator{} uses 50 denoising steps in total. We use $\alpha=15$ steps as our default setting.
  }
  \label{tab:alpha_ablation_results}
\end{table}

\paragraph{Number of denoising steps with layout guidance.}
The number of denoising steps with layout guidance (\ie{}, with the gated self-attention layer activated in each transformer block of the diffusion UNet),
denoted as $\alpha$, is a crucial hyper-parameter in \generator{}. A larger  $\alpha$ value indicates stronger layout control.
\Cref{tab:alpha_ablation_results} presents an ablation study using varying $\alpha$ values.
A smaller $\alpha$ value enhances object generation, while a larger value improves count performance.
This observation aligns with intuition: rigorous layout control more effectively prevents the generation of extraneous objects in the background but may detract from the visual realism of the generated objects, which is also observed in~\citep{li2023gligen, Lin2023VideoDirectorGPT}.
We set the default value for $\alpha$ as 10 steps, as it ensures a good balance of accuracy for objects and counts while achieving optimal performance in depicting object relationships.

\begin{table}[h]
  \centering
    \resizebox{0.7\columnwidth}{!}{
  \begin{tabular}{lccccccccccc|c}
    \toprule

    \multirow{2}{*}{Updated parameters} & \multicolumn{1}{c}{VPEval} & \multicolumn{1}{c}{Captioning} & \multicolumn{1}{c}{CLIPScore} \\
    \cmidrule(lr){2-2} \cmidrule(lr){3-3}\cmidrule(lr){4-4}
    
     & Overall & BERTScore & Img-Txt \\
    \midrule
    None & 68.5 & 88.9 & 29.3 \\
    GatedSA Layers only & 67.4 & 88.9 & 30.0 \\
    All Layers (default) & \textbf{71.2} & \textbf{89.4} & \textbf{32.1} \\
    \bottomrule
  \end{tabular}
  }
  \caption{
  Ablation of fine-tuning different layers of \generator{}. We use the fully fine-tuned model as our default setting. \textit{GatedSA: Gated Self-Attention.}
  }
  \label{tab:full_ft_ablation_results}
\end{table}

\begin{table}[h]
  \centering
    \resizebox{0.8\columnwidth}{!}{
  \begin{tabular}{lccccccccccc|c}
    \toprule

    \multirow{2}{*}{Diagram Plan Source} & \multicolumn{1}{c}{VPEval} & \multicolumn{1}{c}{Captioning} & \multicolumn{1}{c}{CLIPScore} \\
    \cmidrule(lr){2-2} \cmidrule(lr){3-3}\cmidrule(lr){4-4}
    
     & Overall & BERTScore & Img-Txt \\
    \midrule
    Ground-truth + \generator{} (oracle) & \textbf{81.9} & \textbf{89.8} & \textbf{32.3} \\
    GPT-4 + \generator{} & 71.2 & 89.4 & 32.1 \\
    \bottomrule
  \end{tabular}
  }
  \caption{
  Ablation of using ground-truth plans from \dataset{} (\eg{}, oracle performance) instead of GPT-4.
  }
  \label{tab:gt_plans}
\end{table}

\paragraph{Fine-tuning: all layers vs. layout layers.}
In \Cref{tab:full_ft_ablation_results}, we present an ablation study comparing the fine-tuning of only the layout control layers in \generator{} with fine-tuning of the entire model, including the Stable Diffusion backbone.
Full fine-tuning enhances performance and improves the visual quality of the diagrams. 
Therefore, we employ the fully fine-tuned version as our default model for all subsequent experiments.

\paragraph{Using ground-truth \plan{}s.}
We experiment with generating diagrams using ground-truth \plan{}s from \dataset{}. Doing this allows us to measure the upper bound of \generator{} and see how much room our stage 1 has for improvement.
\Cref{tab:gt_plans} shows that using ground-truth plans does indeed do better than GPT-4 and that \generator{} is able to perform better when using ground-truth plans. However, it is interesting to know that using GPT-4 performs very closely to the oracle score.

\section{Limitations}
\label{appendix:limitations}
Our framework can benefit many educational applications, such as
presentation/paper creation, and human-in-the-loop diagram generation/modification. However,
akin to other text-to-diagram/text-to-image generation frameworks,
our framework can also make some errors and
be utilized for potentially harmful purposes (\eg{}, creating false information or misleading diagrams),
and thus should be used with caution in real-world applications (with human supervision, \eg{}, as described in
\cref{subsec:additional_analysis_results}
human-in-the-loop \plan{} editing).
Also, generating a \plan{} using the strongest LLM APIs can be costly,
similar to other recent LLM-based frameworks.
We hope that advances in quantization/distillation
and open-source models will continue to lower the inference cost of LLMs.
Lastly, \generator{} is based on the pretrained weights of GLIGEN and \sdonefour{}.
Therefore, we face similar limitations to these models, including deviations related to the distribution of training datasets,
imperfect generation quality,
and only understanding the English corpus.

\end{document}